\documentclass[10pt,journal,compsoc]{IEEEtran}
\ifCLASSOPTIONcompsoc
  \usepackage[nocompress]{cite}
\else
  \usepackage{cite}
\fi
\ifCLASSINFOpdf
\else
\fi
\usepackage{times}
\usepackage{epsfig}
\usepackage{graphicx}
\usepackage{amsmath}
\usepackage{amssymb}
\usepackage{xcolor}
\usepackage{multirow}
\usepackage{subfigure}
\usepackage{booktabs}
\usepackage{cite}
\usepackage{color, colortbl}
\def\ie{\emph{i.e.}} 

\def\eg{\emph{e.g.}}

\definecolor{myred}{RGB}{234,67,53}
\definecolor{myblue}{RGB}{66,133,244}
\usepackage{hyperref}

\begin{document}
%
\title{
GaitGL: Learning Discriminative Global-Local Feature Representations for Gait Recognition
}
%
%
%
%
\author{Beibei Lin,
        Shunli Zhang,
        Ming Wang,
        Lincheng Li,
        and~Xin~Yu

\IEEEcompsocitemizethanks{
\IEEEcompsocthanksitem Beibei Lin, Shunli Zhang and Ming Wang were with the School of Software Engineering, Beijing Jiaotong University, Beijing, China.
\IEEEcompsocthanksitem Lincheng Li was with the Fuxi Lab, NetEase, Hanzhou, China.
\IEEEcompsocthanksitem Xin Yu was with University of Technology Sydney, Australia.
\IEEEcompsocthanksitem Shunli Zhang (slzhang@bjtu.edu.cn) is the corresponding author.
}
}
\IEEEtitleabstractindextext{%
\begin{abstract}

Gait is one of the most important biometric modalities that can be used to identify a person from walking postures. Existing gait recognition methods either directly establish Global Feature Representation (GFR) from original gait sequences or generate Local Feature Representation (LFR) from several local parts.
However, GFR tends to neglect local details of human postures as the receptive fields become larger in the deeper network layers. Although LFR allows the network to focus on the detailed posture information of each local region, it neglects the relations among different local parts and thus only exploits limited local information of several specific regions.
To solve these issues, we propose a global-local based gait recognition network, named GaitGL, to generate more discriminative feature representations.
To be specific, a novel Global and Local Convolutional Layer (GLCL) is developed to take full advantage of both global visual information and local region details in each layer.
GLCL is a dual-branch structure that consists of a GFR extractor and a mask-based LFR extractor. GFR extractor aims to extract contextual information, \eg, the relationship among various body parts, and the mask-based LFR extractor is presented to exploit the detailed posture changes of local regions. 
In addition, we introduce a novel mask-based strategy to improve the local feature extraction capability. Specifically, we design pairs of complementary masks to randomly occlude feature maps, and then train our mask-based LFR extractor on various occluded feature maps. In this manner, the LFR extractor will learn to fully exploit local information.
Extensive experiments on four popular used datasets, CASIA-B, OU-MVLP, GREW and Gait3D, demonstrate that GaitGL achieves better performance than state-of-the-art gait recognition methods. The average rank-1 accuracy on CASIA-B, OU-MVLP, GREW and Gait3D is 93.6\%, 98.7\%, 68.0\% and 63.8\%, respectively, significantly outperforming the competing methods.
The proposed method has won the first prize in two competitions: Human Identification at a Distance (HID) 2020 \cite{yu2020hid} and HID 2021 \cite{yu2021hid}.
The source code has been released at \url{https://github.com/bb12346/GaitGL}.

\end{abstract}

}

\maketitle

\IEEEdisplaynontitleabstractindextext

%
\IEEEpeerreviewmaketitle

\IEEEraisesectionheading{\section{Introduction}\label{sec:introduction}}

\IEEEPARstart{G}{ait} recognition is one of the most popular biometric technologies, which can be used for human identification at distance. Since walking postures of individuals can be collected from remote cameras, gait recognition has been widely used in many domains, such as access monitoring and intelligent security. However, different from other biometric modalities, such as face, fingerprint and iris, human gaits are usually captured in complex conditions, \eg, view changes, different wearing conditions and low-resolution \cite{connor2018biometric, liao2017pose, yu2006framework, shen2022comprehensive, sepas2022deep, song2022casia, zhang2020learning}. Hence, obtaining robust gait feature representations under diverse complex conditions is still a challenging task.

Existing gait recognition methods propose different CNN-based frameworks to generate discriminative feature representations. The feature representations can be roughly divided into two types: Global Feature Representation (GFR) and Local Feature Representation (LFR). In general, GFR-based methods directly extract gait features from original sequences or templates \cite{chao2019gaitset,shiraga2016geinet,zhang2016siamese,lin2020gait}. In contrast, LFR-based methods first partition human gaits into several fixed local regions and then use CNN to extract local features from each local region \cite{fan2020gaitpart,zhang2019cross}.

However, the aforementioned methods use only global or local gait information for human identification, thus limiting the recognition accuracy. 
More specifically, GFR-based methods are easier to neglect the posture details in deep convolutional layers because the receptive fields will become larger in the deeper network layers and tend to capture global contextual relations. To solve this problem, LFR-based methods partition the human bodies into several local parts and then extract local gait features from each local part. In this manner, the receptive fields in the deeper network layers will be restricted to a local region and thus can better exploit detail information. Nevertheless, LFR-based methods neglect the relations among local regions. 
Moreover, existing partition strategies only adopt a top-down pattern to generate several fixed local regions. We argue that this pattern limits the diversity of local regions and cannot fully exploit local information of a gait sequence.

\begin{figure*}[ht]
\centering
\includegraphics[width=0.90\textwidth]{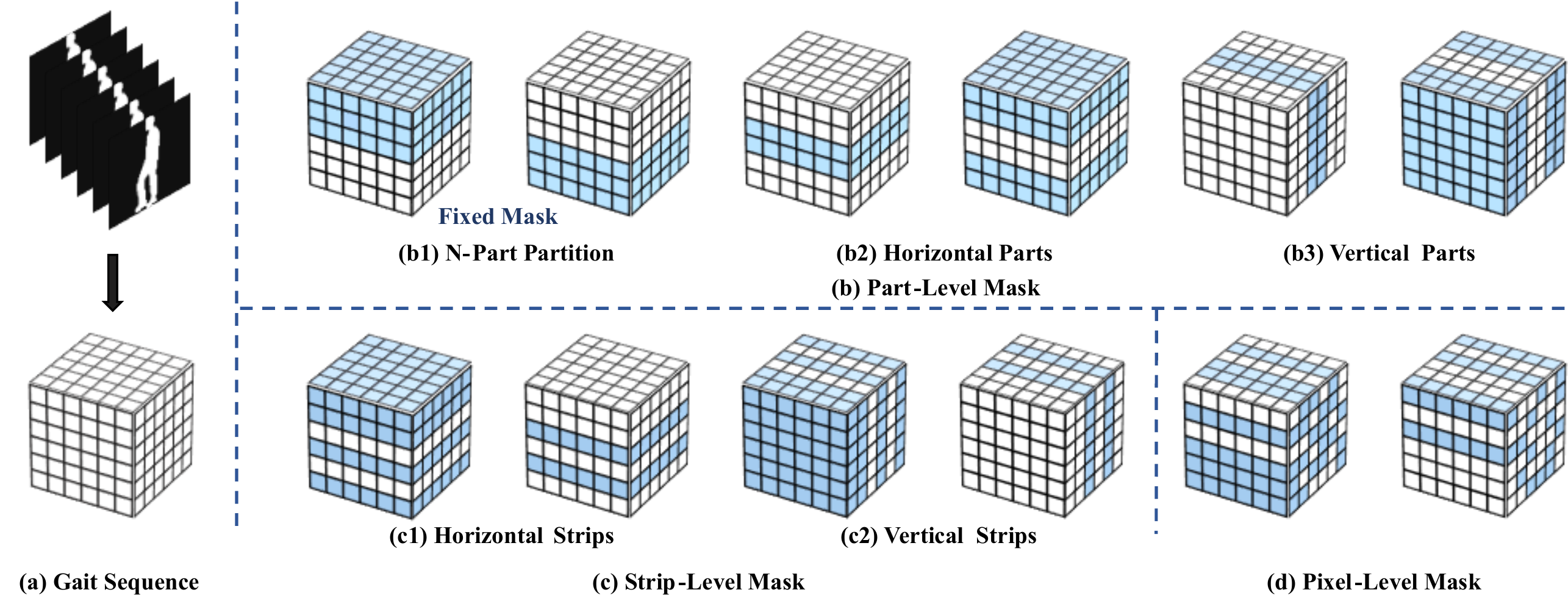}
\caption{Visualization of different mask strategies. (a) represents a gait sequence with six frames. (b), (c) and (d) represent the part-level mask strategy, strip-level mask strategy and pixel-level mask strategy, respectively. The part-level and strip-level mask strategies can be further divided into horizontal and vertical masks. Note that the traditional partition strategy can be viewed as a specific part-level mask. The white block defines original image pixels, while the blue block represents the pixels occluded.
}
\label{Fig_diff_region}
\end{figure*}

To address these problems, in this paper, we propose a novel gait recognition framework, dubbed GaitGL, to learn discriminative representations. GaitGL is built on a novel Global and Local Convolutional Layer (GLCL), which can be used to obtain more comprehensive representations by taking advantage of both global and local information of gait feature maps. GLCL is a dual-branch structure, including a GFR extractor and a mask-based LFR extractor. The GFR extractor aims to extract global contextual information, \eg, the relationships of different parts, while the mask-based LFR extractor is proposed to capture local posture details. 
Unlike traditional partition strategies that only exploit the information of several specified regions, we propose a new mask-based partition strategy to arbitrarily generate local gait regions. 
As shown in Fig. \ref{Fig_diff_region}(b)(c)(d), the proposed mask-based partition strategy randomly generates pairs of complementary masks to occlude gait sequences. Then, the occluded gait sequences can be used to train the LFR extractor. Since the pairs of occluded sequences contain various combinations, the LFR extractor can be fully trained by utilizing various occluded sequences.
Note that the traditional N-part partition can be viewed as a special part-level partition, as illustrated in Fig. \ref{Fig_diff_region}(b1). 
Moreover, we investigate the mask-based partitions in different levels.
\textbf{(1) Part-level.} As show in Fig. \ref{Fig_diff_region}(b), the part-level partition randomly drops a horizontal or vertical part.
\textbf{(2) Strip-level.} As shown in Fig. \ref{Fig_diff_region}(c), the strip-level partition randomly erases multiple inconsecutive horizontal or vertical strips.
\textbf{(3) Pixel-level.} As shown in Fig. \ref{Fig_diff_region}(d), the pixel-level partition randomly occludes a large number of image pixels.

Furthermore, we observe that most state-of-the-art (SOTA) gait recognition methods \cite{chao2019gaitset, chao2021gaitset, fan2020gaitpart, hou2020gait,hou2021set} only adopt spatial pooling layers to downsample the spatial resolution of feature maps at the stage of feature extraction. Nonetheless, they neglect the impact of temporal resolution on the final recognition performance. Intuitively, adjacent frames of a gait sequence are similar and thus contain numerous redundant information. 
For instance, the interval of consecutive frames is 0.03 $\sim$ 0.04 seconds, which only changes slightly. Therefore, we reconsider the gait network design and proposed Local Temporal Aggregation (LTA) to replace the spatial pooling layer. To be specific, the proposed LTA can reduce redundant information by aggregating local temporal information. Meanwhile, removing the spatial pooling layer can preserve larger spatial resolution in which more posture details are contained.
Moreover, we discuss some practical 3D CNN-based network design strategies based on our experience in the competitions: Human Identification at a Distance (HID) 2020 and HID2021.

The main contributions of this paper are four-fold.

(1) We propose a novel gait recognition framework, dubbed GaitGL, to generate discriminative gait representations. To the best of our knowledge, this is the first network to take full advantage of the global and local information of gait sequences.

(2) We present a mask-based LFR extractor to generate more comprehensive local representations by using a pair of complementary masks. Meanwhile, we extensively explore mask-based partition strategies at different levels.


(3) We evaluate the proposed method on four popular used datasets: CASIA-B, OUMVLP, GREW and Gait3D. The experimental results demonstrate that our GaitGL outperforms existing SOTA gait methods by 1.7\% on CASIA-B, 2.2\% on OUMVLP, 17.4\% on GREW and 10.6\% on Gait3D, respectively. In particular, our GaitGL also achieves appealing performance in two competitions, HID 2020 \cite{yu2020hid} and HID 2021 \cite{yu2021hid}.

This work is an extension of our previous works \cite{lin2021gait,lin2021gaitmask}. 
To be specific, we improve our prior works in the following three aspects:
(1) We carry out new experiments to evaluate our GaitGL on real-world datasets.
(2) We conduct more comprehensive ablation studies to analyze the effectiveness of each key module of our GaitGL.
(3) We carefully investigate the different partition strategies and analyze the impact of these strategies on the LFR extractor.

\begin{figure*}[ht]
\centering
\includegraphics[width=0.9\textwidth]{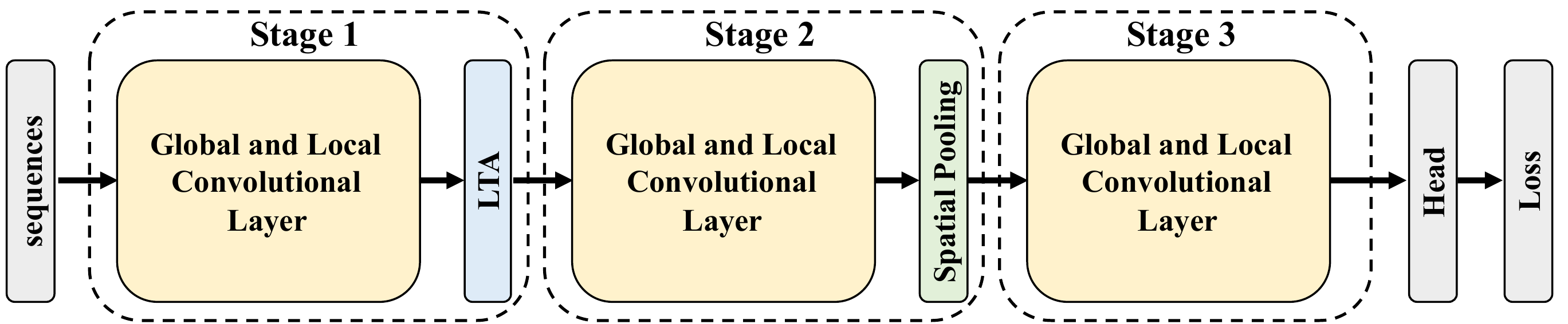}
\caption{Overview of the whole gait recognition framework. The input of our framework is normalized gait sequences. ``LTA'' means Local Temporal Aggregation. ``Head'' indicates Gait Recognition Head, including a spatial feature mapping and a temporal feature mapping. ``Loss'' consists of a cross-entropy loss and a triplet loss. After ``Head'', our network generates discriminative feature representations. During the training stage, the feature representations are used to calculate losses. In the test stage, the feature representations are taken as gait descriptors for evaluation.} 
\label{overview}
\end{figure*}

\section{Related Work}
\subsection{Gait Recognition}
Depending on the data types, gait recognition methods can be roughly classified into three categories, \ie, model-based methods, pose-based methods and silhouette-based methods.

\noindent \textbf{Model-based methods.} The model-based methods \cite{wagg2004automated, li2020end} can utilize much more gait information, \eg, shapes, view angles and postures, to model human motions and generate robust gait feature representations. They generally depend on several images collected from different angles to model effective human motions. However, it is difficult to capture several multi-view gait images in real-world scenes.

\noindent \textbf{Pose-based methods.} The pose-based methods \cite{yam2004automated,teepe2021gaitgraph,an2020performance,liao2020model,an2018improving,sokolova2019pose,li2021end,peng2021learning,jun2020feature} first extract 2D Poses or 3D Poses of the human bodies and then use deep learning-based frameworks to generate feature representations. Since key poses of human bodies are very limited, recent pose-based gait recognition methods cannot generate discriminative feature representations in the real gait dataset. For instance, GREW \cite{zhu2021gait} is one of the largest in-the-wild gait datasets. The accuracy of most pose-based methods on the GREW dataset is less than 5\%. 

\noindent \textbf{Silhouette-based methods.} The silhouette-based methods \cite{zhang2019gait, zhang2020learning} can be roughly divided into three types, \ie, template-based methods, set-based methods and video-based methods. 
In general, the template-based methods \cite{shiraga2016geinet,wu2016comprehensive,zhang2016siamese,li2020gait,wang2020human,liang2022gaitedge} first aggregate all temporal information of a sequence to generate a template, \eg, Gait Energy Image (GEI). Then, CNN-based methods are used to extract gait features from the GEI. For instance, Shiraga et al. \cite{shiraga2016geinet} propose a GEINet framework that consists of two convolutional layers to generate feature representations.
The set-based methods \cite{chao2019gaitset,chao2021gaitset,fan2020gaitpart,hou2020gait,hou2022gait,hou2021set} take each frame of gait sequences as a unit to extract gait features. 
For example, GaitSet \cite{chao2019gaitset,chao2021gaitset} treats gait sequences as an unordered set and then extracts robust gait features from this set. 
Fan et al. \cite{fan2020gaitpart} first extract frame-level gait features and then model temporal information of the feature maps. 
The video-based methods \cite{thapar2019gait, wolf2016multi, lin2020gait, shen2022gait} directly use 3D CNN to extract spatial-temporal features from gait sequences. For example, Thapar et al. \cite{thapar2019gait} first split gait sequences into several fixed-length clips and then use 3D CNN to extract spatial-temporal gait features from each clip. Next, they utilize the Long Short Term Memory (LSTM) to aggregate the features of all clips. Lin et al. \cite{lin2020gait} propose 3D CNN-based framework to extract multi-temporal-scale gait features.
Compared with other methods, video-based methods can better capture spatial-temporal gait changes in the feature extraction stage. Thus, we also use 3D CNN to design our framework.

\subsection{Global and Local Feature Representations}
Recently, gait feature representations can be roughly divided into two types: Global Feature Representation (GFR) and Local Feature Representation (LFR). 

\noindent \textbf{Global feature representation.} In general, GFR-based methods \cite{chao2019gaitset,chao2021gaitset,shiraga2016geinet,hou2020gait,lin2020gait} take the whole human body as a basic unit to extract gait features. For example, Chao et al. \cite{chao2019gaitset, chao2021gaitset} first use 2D CNN to directly extract spatial features from each frame. Then, they use Set Pooling (SP) and Horizontal Pyramid Mapping (HPM) operations to generate robust feature representations. Shiraga et al. \cite{shiraga2016geinet} first generate the GEI by integrating all temporal information of whole gait sequences. Next, they utilize 2D CNN to extract gait features from the generated GEI. Lin et al. \cite{lin2020gait} directly use 3D CNN to generate spatial-temporal feature representations from whole gait sequences. However, the aforementioned methods neglect the local details of human postures as the receptive
fields become larger in the deeper network.

\noindent \textbf{Local feature representation.} LFR-based methods \cite{fan2020gaitpart,zhang2019cross} are presented to better utilize the local postures in deep convolutional layers. To be specific, LFR-based methods partition the human body into several parts and then extract gait features from each local part. For instance, Rokanujjaman et al.~\cite{rokanujjaman2012effective} first partition the human bodies into five parts and then select the most effective three parts for recognition. Zhang et al.~\cite{zhang2019cross} split human gaits into four fixed-length local parts. Then, they use a 2D CNN to extract local gait features from each local part. Finally, LSTM is used to integrate temporal information of gait sequences. Fan et al. \cite{fan2020gaitpart} propose a new convolution layer, named the focal convolution layer, to capture detailed features in each layer.
Nonetheless, the aforementioned methods only exploit local features of gait sequences to represent the human gait, thus limiting the effectiveness of feature representations. To be specific, they ignore the relations among different local regions because the feature of each local region is extracted independently. On the other hand, we observe that these LFR-based methods only adopt a top-to-down partition strategy to split the human body. 
As shown in Fig. \ref{Fig_diff_region}(b1), when LFR employs this strategy to extract features from some parts of local regions, it may often lose some  details of other regions.

To solve the issues in both GFR-based and LFR-based methods, in this paper, we propose a novel Global and Local Convolutional Layer (GLCL) to generate discriminative feature representation.
The proposed GLCL takes advantage of both global relations and local detailed information. Furthermore, we present a new partition strategy, dubbed mask-based partition, to better exploit local detailed information, which is shown in Fig. \ref{Fig_diff_region}(b)(c)(d). More specifically, the mask-based partition randomly generates pairs of complementary masks to occlude gait sequences. During the training stage, various occluded feature maps that contain different local detailed information can be fed into the LFR extractor to  improve local feature extraction ability.

\section{Proposed Method} \label{method}
In this section, we first introduce the pipeline of the proposed method, which can be used to generate discriminative gait feature representation. Then, we describe some key modules, including the Local Temporal Aggregation (LTA) and the Global and Local Convolutional Layer (GLCL). Finally, we explain the details of feature mapping, training and test. 

\subsection{Overall Architecture}
The overview of our GaitGL is shown in Fig. \ref{overview}. The gait sequences which are taken as the input of the proposed framework are fed into the global and local convolutional layer to extract features. Next, a pooling layer is used to downsample the resolution of feature maps. The global and local convolutional layer and the pooling layer are defined as ``Stage 1''. We repeat the ``Stage 1'' twice as ``Stage 2'' and ``Stage 3''. Note that we remove the pooling layer of ``Stage 3''. Previous SOTA methods \cite{chao2019gaitset, fan2020gaitpart,chao2021gaitset,hou2020gait,hou2021set} only use spatial pooling layers to downsample the spatial resolution of feature maps. In this paper, we propose a Local Temporal Aggregation (LTA) to trade off temporal and spatial information of feature maps. To be specific, the LTA is used to replace the first spatial pooling layer (in ``Stage 1'').

Assume that the feature map $X_{i}\in \mathbb{R}^{C_{i} \times T_{i} \times H_{i} \times W_{i}}$ is taken as the network input, where $C_i$ is the number of channels, $T_{i}$ is the length of gait sequences and ($H_{i}$,$W_{i}$) is the image size of each frame. First, the feature map $X_{i}$ is fed into the ``Stage 1'' and the output of the ``Stage 1'' is the feature map $X_{s_1} \in \mathbb{R}^{C_{s_1} \times T_{o} \times H_{i}  \times W_{i}}$, where $T_{o}$ is the number of frames of the feature map $X_{s_1}$. Then, the ``Stage 2'' is applied into the feature map $X_{s_1}$ to generate the feature map $X_{s_2} \in \mathbb{R}^{C_{s_2} \times T_{o} \times \frac{H_{i}}{2} \times \frac{W_{i}}{2}}$. After the ``Stage 3'', the final feature map $X_{s_3} \in \mathbb{R}^{C_{s_3} \times T_{o} \times \frac{H_{i}}{2} \times \frac{W_{i}}{2}}$ is generated. Finally, the final feature maps are fed into a gait recognition head to generate discriminative feature representations \cite{fan2020gaitpart,lin2021gait,lin2021gaitmask}.
During the training stage, a combined loss function, which consists of both the triplet loss and cross-entropy loss, is used to train the proposed model \cite{chao2021gaitset,lin2021gait}.

\begin{figure}[t]
\centering
\includegraphics[width=0.49\textwidth]{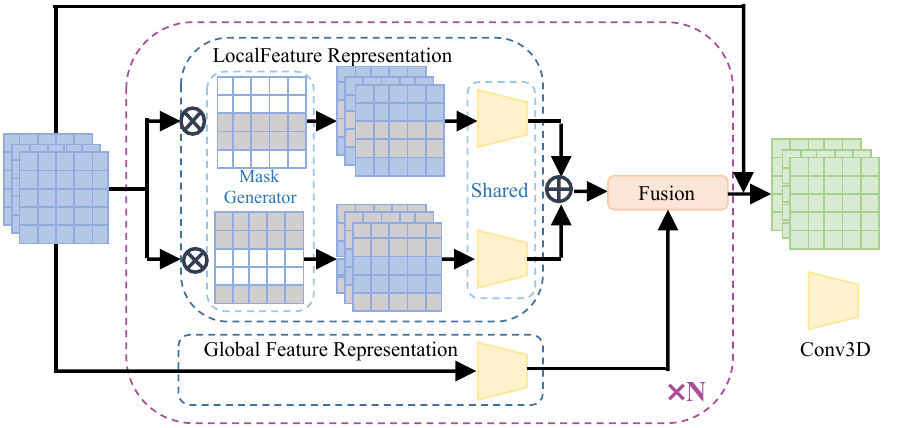}
\vspace*{-1.5em}
\caption{Architectures of Global and Local Convolutional Layer. $\oplus$ means the element-wise addition, $\otimes$ means the element-wise multiplication and ``Conv3D'' means a 3D convolution operation. $N$ represents the number of times using both the local feature extractor and global feature extractor. The parameters of all convolutions in the local feature extractor are shared. ``Mask Generator'' means generating different masks by using the proposed mask strategies.
}
\label{Fig_GLCL}
\end{figure}

\subsection{Local Temporal Aggregation}
Previous works \cite{chao2019gaitset,fan2020gaitpart,hou2020gait,chao2021gaitset} only adopt spatial pooling layers to downsample the spatial resolution of feature maps. Although higher spatial resolution can preserve much more posture details of the human body, using high-resolution images to train the recognition network is not feasible due to the memory limit. However, we observe that adjacent frames of gait sequences are generally collected in less than 0.04 seconds. In other words, they contain numerous redundant temporal information. Based on above fact, we propose the Local Temporal Aggregation (LTA) operation to trade off the extraction of spatial and temporal information. As shown in Fig. \ref{overview}, we use a LAT module to replace the second spatial pooling layer. LTA is used to aggregate local temporal information of gait sequences which can remove redundant temporal information. Meanwhile, removing a spatial pooling layer can achieve higher spatial resolution.

Assume that $X_{l}\in \mathbb{R}^{C_{l} \times T_{l} \times H_{l} \times W_{l}}$ is the input feature map of LTA, where $C_{l}$ is the number of input channels, $T_{l}$ is the length of the gait sequence and ($H_{l}$,$W_{l}$) is the image size of each frame. LTA can be formulated as
\begin{equation}
X_{LTA} =  F_{a\times a\times a}^{b \times1\times1}(X_{in}),
\end{equation}
where $F_{a\times a\times a}^{b \times1\times1}(\cdot)$ denotes a 3D convolution with kernel size $a$ and temporal stride $b$. $X_{LTA} \in \mathbb{R}^{C_{l} \times T_{o} \times H_{l} \times W_{l}}$ is the output of LTA, in which $T_{o} = \lfloor  \frac{T_{l} - a}{b} \rfloor + 1$. After LTA, the image size of input sequences remains the same as before.

\subsection{Global and Local Convolutional Layer}
At present, most gait recognition works utilize either Global Feature Representation (GFR) or Local Feature Representation (LFR) to portray human gait. 
Although LFR pays more attention to the details of each local part, it neglects the relations among different parts. In contrast, GFR is hard to capture local details of feature maps in a deep layer since the receptive fields become larger in the deep layer.
To solve these problems, we propose a novel Global and Local Convolutional Layer (GLCL) to take full advantage of both global and local information of gait sequences. As shown in Fig. \ref{Fig_GLCL}, GLCL includes two extractors: a GFR extractor and a LFR extractor. 

Assume that the input of GLCL is $X_{in} \in \mathbb{R}^{C_{in} \times T_{in} \times H_{in} \times W_{in}}$, where $C_{in}$ is the number of channels, $T_{in}$ is the length of feature maps and ($H_{in}$,$W_{in}$) is the image size of each frame. GLCL can be represented as
\begin{equation}
Y_{GLCL-A} = F_{GFR}(X_{in}) + F_{LFR}(X_{in}),
\end{equation}
\begin{equation}
Y_{GLCL-B} = Concat\begin{Bmatrix}
F_{GFR}(X_{in})\\
F_{LFR}(X_{in})
\end{Bmatrix},
\end{equation}
where $F_{GFR}(\cdot)$ and $F_{LFR}(\cdot)$ denote the GFR extractor and the LFR extractor, respectively. $Y_{GLCL-A}\in \mathbb{R}^{C_{out} \times T_{in} \times H_{in} \times W_{in}}$ and $Y_{GLCL-B}\in \mathbb{R}^{C_{out} \times T_{in} \times 2*H_{in} \times W_{in}}$ are two types of the GLCL output.

\subsubsection{Global Feature Representation}

The GFR extractor is shown in Fig. \ref{Fig_GLCL}. It uses a 3D convolution to directly extract gait features from the input feature map, denoted as
\begin{equation}
F_{GFR}(X_{in}) = F_{k\times k\times k}(X_{in}),
\end{equation} 
where $F_{k\times k\times k}(\cdot)$ denotes the 3D convolution operation with kernel size $k$. 

\subsubsection{Local Feature Representation}
\label{sec_LFR}
Previous works \cite{fan2020gaitpart,zhang2019cross} usually partition the human gait into N-parts and extract gait features from each local part. 
Assume that $X_{in} = \left \{ X_{in}^{l} | l=1,...,n \right \}$, where $n$ is the number of partitions and $ X_{in}^{l}  \in \mathbb{R}^{C_{in} \times T_{in} \times \frac{H_{in}}{n} \times W_{in}}$ corresponds to the $l$-th local gait part, the traditional LFR extractor can be formulated as

\begin{equation}
\label{Equ_tLFR}
F_{LFR}^{Tra}(X_{in}) = Concat\begin{Bmatrix}
F_{k\times k\times k}(X_{in}^{1})\\
F_{k\times k\times k}(X_{in}^{2})\\
...\\
F_{k\times k\times k}(X_{in}^{n})
\end{Bmatrix},
\end{equation}
where ``Concat'' means concatenating the feature maps of different parts horizontally. Note that all convolution operations in Eqn.\ref{Equ_tLFR} are shared. 

However, the N-part partition only focuses on the information of a few specific local regions. It cannot fully exploit detailed postures of gait feature maps. Therefore, we further propose a mask-based LFR extractor that is shown in Fig. \ref{Fig_GLCL}.
The mask-based LFR extractor first generates a pair of complementary masks $M_{p} \in \mathbb{R}^{H_{in} \times W_{in}}$ and $M_{q} \in \mathbb{R}^{H_{in} \times W_{in}}$. All elements of $M_{p}$ are initialized to 0, while all elements of $M_{q}$ are initialized to 1, respectively.
Then, we randomly drop some regions of the mask $M_{q}$. Meanwhile, the corresponding regions in the mask $M_{p}$ are set to 1. Therefore, a new pair of complementary masks $\widetilde{M}_{p} \in \mathbb{R}^{H_{in} \times W_{in}}$ and $\widetilde{M}_{q} \in \mathbb{R}^{H_{in} \times W_{in}}$ can be generated. Then, the generated feature maps $\widetilde{M}_{p}$ and $\widetilde{M}_{q}$ can be used to occlude the input feature maps. The mask-based LFR extractor can be represented as
\begin{equation}
\begin{aligned}
F_{LFR}^{Mask}(X_{in}) = F_{k\times k\times k}( \sum^{C_{in}}_{y=1} \sum^{T_{in}}_{z=1}  (  X_{in}^{y,z} \otimes \widetilde{M}_{p}) ) + \\
F_{k\times k\times k}(\sum^{C_{in}}_{y=1} \sum^{T_{in}}_{z=1}  ( X_{in}^{y,z} \otimes \widetilde{M}_{q}) ),
\end{aligned}
\end{equation}
where $F_{LFR}^{Mask}(\cdot)$ denotes the mask-based LFR extractor. $F_{k\times k\times k}$ represents the 3D convolution operation with kernel size k. $X_{in}^{y,z}$ denote the $y$-th channel and the $z$-th frame of the feature map $X_{in}$. ``$\otimes$'' means element-wise multiplication.

According to the levels of occlusion, we propose three different mask strategies: Part-level, Strip-level and Pixel-level. 

\noindent \textbf{Part-level mask.} 
As shown in Fig. \ref{Fig_diff_region}, the part-level mask strategy randomly drops a local part of feature maps, where each part denotes continuous rows or columns. To be specific, it includes two types of masks, i.e. horizontal masks at the part-level and vertical masks at the part-level.

\emph{(1) Horizontal masks at the part level.} Assume that $M_{p} = \left \{ h_{p}^{i}|i = 1,2,...,H_{in} \right \} $ where $h_{p}^{i}\in \mathbb{R}^{ 1 \times W_{in}}$ is the $i$-th row of the feature map $M_{p}$, and
$M_{q} = \left \{ h_{q}^{j}|j = 1,2,...,H_{in} \right \} $ where $h_{q}^{j}\in \mathbb{R}^{ 1 \times W_{in}}$ is the $j$-th row of the feature map $M_{q}$. 
The horizontal mask at the part-level first randomly selects a horizontal interval $(r, r+\left \lfloor d \times H_{in} \right \rfloor)$, where $r$ denotes the $r$-th row of the feature map $M_{q}$ and $d$ means the dropping ratio. Then, the value of $\left \{ h_{p}^{r},...,h_{p}^{r+\left \lfloor d \times H_{in} \right \rfloor} \right \}$ in the mask $M_{p}$ is set to 1, as a new mask $\widetilde{M}^{hpart}_{p} \in \mathbb{R}^{H_{in} \times W_{in}}$, while the value of $\left \{ h_{q}^{r},...,h_{q}^{r+\left \lfloor d \times H_{in} \right \rfloor} \right \}$ in the mask $M_{q}$ is set to 0, as a new mask $\widetilde{M}^{hpart}_{q} \in \mathbb{R}^{H_{in} \times W_{in}}$.

\emph{(2) Vertical masks at the part level.} Assume that $M_{p} = \left \{ v_{p}^{i}|i = 1,2,...,W_{in} \right \} $ where $v_{p}^{i}\in \mathbb{R}^{ H_{in} \times 1}$ is the $i$-th column of the feature map $M_{p}$, and
$M_{q} = \left \{ v_{q}^{j}|j = 1,2,...,W_{in} \right \} $ where $v_{q}^{j}\in \mathbb{R}^{ H_{in} \times 1}$ is the $j$-th column of the feature map $M_{q}$. The vertical mask at the part-level first randomly selects a vertical interval $(r, r+\left \lfloor d \times W_{in} \right \rfloor)$,  where $r$ denotes the $r$-th column of the feature map $M_{q}$ and $d$ means the dropping ratio. Then, the value of $\left \{ v_{p}^{r},...,v_{p}^{r+\left \lfloor d \times W_{in} \right \rfloor} \right \}$ in the mask $M_{p}$ is set to 1, as a new mask $\widetilde{M}^{vpart}_{p} \in \mathbb{R}^{H_{in} \times W_{in}}$, while the value of $\left \{ v_{q}^{r},...,v_{q}^{r+\left \lfloor d \times W_{in} \right \rfloor} \right \}$ in the mask $M_{q}$ is set to 0, as a new mask $\widetilde{M}^{vpart}_{q} \in \mathbb{R}^{H_{in} \times W_{in}}$.

\noindent \textbf{Strip-level mask.} Fig. \ref{Fig_diff_region} shows the strip-level mask strategy randomly drops a few strips of feature maps, where each strip denotes a row or column. It also includes two types of masks, horizontal masks at the strip level and vertical masks at the strip level.

\noindent \emph{(1) Horizontal masks at the strip level.} 
The horizontal mask at the strip-level first randomly selects a subset $\mathbb{S}_h$ of $M_{p} = \left \{ h_{p}^{i}|i = 1,2,...,H_{in} \right \} $. The number of strips in $\mathbb{S}_h$ is $d \times H_{in}$, where $d$ means the dropping ratio. Then, the value of $\mathbb{S}_h$ in the mask $M_{p}$ is set to 1, as a new mask $\widetilde{M}^{hstrip}_{p} \in \mathbb{R}^{H_{in} \times W_{in}}$. Meanwhile, the corresponding strips of the mask $M_{q}$ are set to 0, as a new mask $\widetilde{M}^{hstrip}_{q} \in \mathbb{R}^{H_{in} \times W_{in}}$.

\noindent \emph{(2) Vertical masks at the strip level.} The vertical mask at the strip-level also randomly selects a subset $\mathbb{S}_v$ of $M_{p} = \left \{ v_{p}^{j}|j = 1,2,...,W_{in} \right \} $, where the number of strips in $\mathbb{S}_v$ is $d \times W_{in}$. Then, the value of $\mathbb{S}_v$ in the mask $M_{p}$ is set to 1, as a new mask $\widetilde{M}^{vstrip}_{p} \in \mathbb{R}^{H_{in} \times W_{in}}$. On the other hand, the corresponding strips of the mask $M_{q}$ are set to 0, as a new mask $\widetilde{M}^{vstrip}_{q} \in \mathbb{R}^{H_{in} \times W_{in}}$.

\noindent \textbf{Pixel-level mask.} As shown in Fig. \ref{Fig_diff_region}, the pixel-level mask strategy randomly drops some pixels of gait sequences. 
Assume that $M_{p} = \left \{ hv_{p}^{(i,j)}|i = 1,...,H_{in}, j = 1,...,W_{in} \right \} $ where $hv_{p}^{(i,j)}\in \mathbb{R}^{ 1 \times 1}$ is the ($i$-th, $j$-th) pixel of the feature map $M_{p}$, and $M_{q} = \left \{ hv_{q}^{(i,j)}|i = 1,...,H_{in}, j = 1,...,W_{in} \right \} $ where $hv_{q}^{(i,j)}\in \mathbb{R}^{ 1 \times 1}$ is the ($i$-th, $j$-th) pixel of the feature map $M_{q}$. The pixel-level mask first randomly selects a subset $\mathbb{U}$ of $M_{p}$, where the number of the pixels in $\mathbb{U}$ is $d \times (H_{in} \times W_{in})$, and $d$ means the dropping ratio. Then, the value of $\mathbb{U}$ in the mask $M_{p}$ is set to 1, as a new mask $\widetilde{M}^{pixel}_{p} \in \mathbb{R}^{H_{in} \times W_{in}}$, while the corresponding pixels of the mask $M_{q}$ are set to 0, as a new mask $\widetilde{M}^{pixel}_{q} \in \mathbb{R}^{H_{in} \times W_{in}}$.

\subsection{Gait Recognition Head}
\label{Sec_FP}
After feature extraction, a gait recognition head is presented to map feature maps into vectors for representation.
The gait recognition head includes a temporal feature mapping module and a spatial feature mapping module.

\noindent \textbf{Temporal feature mapping.} Since the length of the input gait sequences is different in the test stage, we introduce a max-pooling layer to integrate all temporal information of a gait sequence \cite{lin2020gait}.
Assume that $X_{fm}\in \mathbb{R}^{C_{fm} \times T_{fm} \times H_{fm} \times W_{fm}}$ is the input of the temporal feature mapping, where $C_{fm}$ is the number of input channels, $T_{fm}$ is the length of feature maps and ($H_{fm}$,$W_{fm}$) is the image size of feature maps, the temporal feature mapping can be formulated as
\begin{equation} \label{Equ_tfm}
  Y_{TFM} =  F_{Max}^{1 \times T_{fm} \times 1 \times 1}(X_{fm}),
\end{equation}
where $F_{Max}^{1 \times T_{fm} \times 1 \times 1}(\cdot)$ means max-pooling layer with kernel size $(1 \times T_{fm} \times 1 \times 1)$. $Y_{TFM} \in \mathbb{R}^{C_{fm} \times 1 \times H_{fm} \times W_{fm}}$ is the output of temporal feature mapping.

\noindent \textbf{Spatial feature mapping.} To preserve more spatial information, previous works usually first split the feature maps into multiple horizontal strips, and then aggregate the vertical information of each strip by using average and max pooling \cite{chao2019gaitset,fan2020gaitpart}. The previous spatial feature mapping can be designed as
\begin{equation}
  Y_{SFM}^{MA} = \alpha F_{Max}^{1\times1\times1\times W_{fm}}(Y_{TFM}) + \beta F_{Avg}^{1\times1\times1\times W_{fm}}(Y_{TFM}),
  \label{Equ_MA}
\end{equation}
where $Y_{SFM} \in \mathbb{R}^{C_{fm} \times 1 \times H_{fm} \times 1}$ is the output of spatial feature mapping. $ \alpha$ and $ \beta$ can be 0 or 1. ``$\alpha = 0$ and $\beta = 1$'' represents using only the average pooling layer, while ``$\alpha = 1$ and $\beta = 0$'' means using only the max pooling layer.
However, this operation directly uses two pooling operations to ensemble gait information and do not consider the effectiveness of different pooling operation on different datasets.

Therefore, we further introduce the Generalized-Mean pooling (GeM) \cite{radenovic2018fine} to aggregate the vertical information of feature maps adaptively. The GeM pooling layer can be formulated as
\begin{equation}
  Y_{SFM}^{GeM} = F_{GeM}(Y_{TFM}),
  \label{Equ_SFM}
\end{equation}
\begin{equation}
  F_{GeM}(Y_{TFM}) = (F_{Avg}^{1\times1\times1\times W_{fm}}( (Y_{TFM})^{p}))^{\frac{1}{p}},
  \label{Equ_GeM}
\end{equation}
where $Y_{SFM}^{GeM} \in \mathbb{R}^{C_{fm} \times 1 \times H_{fm} \times 1}$ is the output of GeM. The parameter $p$ can be optimized during the training stage.
Specifically, if $p=1$, $F_{GeM}(\cdot)$ is equal to $F_{Avg}^{1\times1\times1\times W_{fm}}(\cdot)$, and if $p \rightarrow \infty$, $F_{GeM}(\cdot)$ is close to $F_{Max}^{1\times1\times1\times W_{fm}}(\cdot)$. 

After spatial feature mapping, we introduce multiple separate Fully Connected (FC) layers to map gait features into a more discriminative representation space \cite{chao2019gaitset,chao2021gaitset},
\begin{equation}
Y_{STFM} =  {F}_{sfc}(Y_{SFM}),
\end{equation}
where $Y_{STFM} \in \mathbb{R}^{C_{stfm} \times 1 \times H_{fm} \times 1}$ is the output of the spatial and temporal feature mappings. $F_{sfc}(\cdot)$ denotes the multiple separate fully connected layers.

\subsection{Training and Test Details}
\label{Traintest}

\noindent \textbf{Training stage.} In this paper, a combined loss, consisting of the triplet loss \cite{hermans2017defense} and cross-entropy loss, is presented to train the proposed network. The triplet loss aims to optimize the inter-class and intra-class distance. Specifically, it increases the inter-class distance and reduces the intra-class distance in the representational space.

After spatial-temporal feature mapping, the feature representation $Y_{STFM} \in \mathbb{R}^{C_{stfm} \times 1 \times H_{fm} \times 1}$ can be obtained. It includes $H_{fm}$ horizontal vectors, where the length of each horizontal vector is $C_{stfm}$. During the training stage, each horizontal vector is fed into the combined loss function to calculate the loss independently. The combined loss function $L_{c}$ can be formulated as
\begin{equation}
L_{c} = L_{tri} + L_{cse},
\end{equation}
where $L_{tri}$ and $L_{cse}$ denote the triplet loss and cross entropy loss, respectively. The triplet loss is defined as 
\begin{equation}
\label{tri}
L_{tri} = [D(Y_{STFM}^{A_i},Y_{STFM}^{B_k}) - D(Y_{STFM}^{A_i},Y_{STFM}^{A_j}) + m]_{+},
\end{equation}
where $A_i$ and $A_j$ are the samples from the class A, while $B_k$ denotes the sample from another class is B. $D(d_i, d_j)$ is the Euclidean distance between $d_i$ and $d_j$. $m$ is the margin of the triplet loss. The operation $[\gamma]_{+}$ is equal to $max(\gamma,0)$. 

To better optimize the triplet loss function, we take the Batch ALL (BA) strategy as the sampling strategy \cite{hermans2017defense,chao2019gaitset,fan2020gaitpart}. Specifically, the number of samples in a batch is $P \times K$, where $P$ denotes the number of subject IDs and $K$ means the number of samples for a subject ID. During the training stage, we set the length of the input sequence to $T$ frames due to the limited memory.

\noindent \textbf{Test stage.} During the test stage, all frames of a gait sequence can be fed into the proposed network to generate the feature representation $Y_{STFM} \in \mathbb{R}^{C_{stfm} \times 1 \times H_{fm} \times 1}$.
In general, the test dataset is divided into two sets for evaluation, i.e. the gallery set and the probe set. The samples from the gallery set are labeled data, while the samples from the probe set are used for prediction. After feature extraction, we calculate the Euclidean distance between the  feature of each sample in the probe set and all features of samples in the gallery set. For a given probe sample,  the label belonging to the gallery sample which has the smallest distance with this probe sample is assigned. Finally, we can calculate the recognition accuracy.


\begin{table*}[ht]
  \scriptsize
  \centering
   \caption{Rank-1 accuracy (\%) on the CASIA-B dataset under all view angles and different conditions, excluding the identical-view case. ``NM'', ``BG'' and ``CL'' mean normal walking, walking with a bag, walking with a coat, respectively.
} 
 \renewcommand\arraystretch{1.2} 
  \resizebox{0.99\textwidth}{!}{

    \begin{tabular}{c|c|c|c|c|c|c|c|c|c|c|c|c|c}
    \toprule
    \multicolumn{2}{c|}{\textbf{Gallery}} & \multicolumn{12}{c}{\textbf{$0^{\circ}$--$180^{\circ}$}} \\
    \hline
    \multicolumn{2}{c|}{\textbf{Probe}} & $0^{\circ}$     & $18^{\circ}$    & $36^{\circ}$    & $54^{\circ}$    & $72^{\circ}$    & $90^{\circ}$    & $108^{\circ}$   & $126^{\circ}$   & $144^{\circ}$   & $162^{\circ}$   & $180^{\circ}$  & Mean \\
    \midrule
  
    \multicolumn{1}{c|}{\multirow{7}[2]{*}{NM}} & GaitSet \cite{chao2019gaitset} & 90.8  & 97.9  & 99.4  & 96.9  & 93.6  & 91.7  & 95.0  & 97.8  & 98.9  & 96.8  & 85.8  & 95.0   \\
\cline{2-14}          & GaitPart \cite{fan2020gaitpart} & 94.1  & 98.6  & 99.3  & 98.5  & 94.0  & 92.3  & 95.9  & 98.4  & 99.2  & 97.8  & 90.4  & 96.2   \\
\cline{2-14}          & MT3D \cite{lin2020gait} & 95.7  & 98.2  & 99.0  & 97.5  & 95.1  & 93.9  & 96.1  & 98.6  & 99.2  & 98.2  & 92.0  & 96.7   \\
\cline{2-14}          & 3D Local \cite{huang20213d} & 96.0  & 99.0  & \textbf{\textcolor{myred}{99.5}}  & \textbf{\textcolor{myred}{98.9}}  & \textbf{\textcolor{myred}{97.1}}  & 94.2  & 96.3  & \textbf{\textcolor{myblue}{99.0}}  & 98.8  & 98.5  & 95.2  & 97.5   \\
\cline{2-14}          & CSTL \cite{huang2021context} & \textbf{\textcolor{myred}{97.2}} & \textbf{\textcolor{myred}{99.0}} & \textbf{\textcolor{myblue}{99.2}} & 98.1  & 96.2  & \textbf{\textcolor{myblue}{95.5}}  & \textbf{\textcolor{myblue}{97.7}}  & 98.7  & 99.2  & \textbf{\textcolor{myblue}{98.9}}  & \textbf{\textcolor{myred}{96.5}} & \textbf{\textcolor{myblue}{97.8}}   \\
\cline{2-14}          & Our preliminary \cite{lin2021gait} & 96.0  & 98.3  & 99.0  & 97.9  & 96.9  & 95.4  & 97.0  & 98.9  & \textbf{\textcolor{myblue}{99.3}}  & 98.8  & 94.0  & 97.4   \\
\cline{2-14}          & Ours  & \textbf{\textcolor{myblue}{96.6}}  & \textbf{\textcolor{myblue}{98.8}}  & 99.1  & \textbf{\textcolor{myblue}{98.1}} & \textbf{\textcolor{myblue}{97.0}} & \textbf{\textcolor{myred}{96.8}} & \textbf{\textcolor{myred}{97.9}} & \textbf{\textcolor{myred}{99.2}} & \textbf{\textcolor{myred}{99.3}} & \textbf{\textcolor{myred}{99.3}} & \textbf{\textcolor{myblue}{95.6}}  & \textbf{\textcolor{myred}{98.0}}  \\
    \midrule
    \midrule
    \multicolumn{1}{c|}{\multirow{7}[2]{*}{BG}} & GaitSet \cite{chao2019gaitset} & 83.8  & 91.2  & 91.8  & 88.8  & 83.3  & 81.0  & 84.1  & 90.0  & 92.2  & 94.4  & 79.0  & 87.2   \\
\cline{2-14}          & GaitPart \cite{fan2020gaitpart} & 89.1  & 94.8  & 96.7  & 95.1  & 88.3  & 84.9  & 89.0  & 93.5  & 96.1  & 93.8  & 85.8  & 91.5   \\
\cline{2-14}          & MT3D \cite{lin2020gait}  & 91.0  & 95.4  & 97.5  & 94.2  & 92.3  & 86.9  & 91.2  & 95.6  & 97.3  & 96.4  & 86.6  & 93.0   \\
\cline{2-14}          & 3D Local \cite{huang20213d} & \textbf{\textcolor{myblue}{92.9}}  & 95.9  & \textbf{\textcolor{myred}{97.8}}  & \textbf{\textcolor{myblue}{96.2}}  & 93.0  & 87.8  & \textbf{\textcolor{myblue}{92.7}}  & 96.3  & 97.9  & \textbf{\textcolor{myred}{98.0}}  & 88.5  & 94.3   \\
\cline{2-14}          & CSTL \cite{huang2021context}  & 91.7  & 96.5  & 97.0  & 95.4  & 90.9  & 88.0  & 91.5  & 95.8  & 97.0  & 95.5  & 90.3  & 93.6   \\
\cline{2-14}          & Our preliminary \cite{lin2021gait} & 92.6  & \textbf{\textcolor{myblue}{96.6}}  & 96.8  & 95.5  & \textbf{\textcolor{myblue}{93.5}}  & \textbf{\textcolor{myblue}{89.3}}  & 92.2  & \textbf{\textcolor{myblue}{96.5}}  & \textbf{\textcolor{myblue}{98.2}}  & 96.9  & \textbf{\textcolor{myblue}{91.5}}  & \textbf{\textcolor{myblue}{94.5}}   \\
\cline{2-14}          & Ours  & \textbf{\textcolor{myred}{93.9}} & \textbf{\textcolor{myred}{97.3}} & \textbf{\textcolor{myblue}{97.6}} & \textbf{\textcolor{myred}{96.2}} & \textbf{\textcolor{myred}{94.7}} & \textbf{\textcolor{myred}{91.0}} & \textbf{\textcolor{myred}{94.4}} & \textbf{\textcolor{myred}{97.2}} & \textbf{\textcolor{myred}{98.6}} & \textbf{\textcolor{myblue}{97.1}} & \textbf{\textcolor{myred}{91.6}} & \textbf{\textcolor{myred}{95.4}}  \\
    \midrule
    \midrule
    \multicolumn{1}{c|}{\multirow{7}[2]{*}{CL}} & GaitSet \cite{chao2019gaitset} & 61.4  & 75.4  & 80.7  & 77.3  & 72.1  & 70.1  & 71.5  & 73.5  & 73.5  & 68.4  & 50.0  & 70.4   \\
\cline{2-14}          & GaitPart \cite{fan2020gaitpart} & 70.7  & 85.5  & 86.9  & 83.3  & 77.1  & 72.5  & 76.9  & 82.2  & 83.8  & 80.2  & 66.5  & 78.7   \\
\cline{2-14}          & MT3D \cite{lin2020gait}  & 76.0  & 87.6  & 89.8  & 85.0  & 81.2  & 75.7  & 81.0  & 84.5  & 85.4  & 82.2  & 68.1  & 81.5   \\
\cline{2-14}          & 3D Local \cite{huang20213d} & \textbf{\textcolor{myblue}{78.2}}  & \textbf{\textcolor{myblue}{90.2}}  & \textbf{\textcolor{myblue}{92.0}}  & 87.1  & 83.0  & 76.8  & 83.1  & 86.6  & 86.8  & 84.1  & 70.9  & 83.7   \\
\cline{2-14}          & CSTL \cite{huang2021context}  & 78.1  & 89.4  & 91.6  & 86.6  & 82.1  & \textbf{\textcolor{myblue}{79.9}}  & 81.8  & 86.3  & \textbf{\textcolor{myblue}{88.7}}  & \textbf{\textcolor{myblue}{86.6}}  & \textbf{\textcolor{myblue}{75.3}}  & \textbf{\textcolor{myblue}{84.2}}   \\
\cline{2-14}          & Our preliminary \cite{lin2021gait} & 76.6  & 90.0  & 90.3  & \textbf{\textcolor{myblue}{87.1}}  & \textbf{\textcolor{myblue}{84.5}}  & 79.0  & \textbf{\textcolor{myblue}{84.1}}  & \textbf{\textcolor{myblue}{87.0}}  & 87.3  & 84.4  & 69.5  & 83.6   \\
\cline{2-14}          & Ours  & \textbf{\textcolor{myred}{82.6}} & \textbf{\textcolor{myred}{92.6}} & \textbf{\textcolor{myred}{94.2}} & \textbf{\textcolor{myred}{91.8}} & \textbf{\textcolor{myred}{86.1}} & \textbf{\textcolor{myred}{81.3}} & \textbf{\textcolor{myred}{87.2}} & \textbf{\textcolor{myred}{90.2}} & \textbf{\textcolor{myred}{90.9}} & \textbf{\textcolor{myred}{88.5}} & \textbf{\textcolor{myred}{75.4}} & \textbf{\textcolor{myred}{87.3}}  \\
    \bottomrule
    \end{tabular}%

  } 

\label{comparision_casia}
\end{table*}%

\section{Experiments}

\subsection{Datasets}

In this paper, we evaluate the proposed GaitGL on four popular datasets, CASIA-B \cite{yu2006framework}, OUMVLP \cite{takemura2018multi}, GREW \cite{takemura2018multi} and Gait3D \cite{zheng2022gait}.

\noindent \textbf{CASIA-B} \cite{yu2006framework} is one of the most popular gait datasets. It contains 124 subjects with three different variations, including view angles, clothing and carrying conditions. For each subject, a total of 11 view angles ($0^{\circ}$-$180^{\circ}$ with interval $18^{\circ}$). 
For each view, 10 groups of videos with three different settings (normal walking (NM), walking with a bag (BG) and walking with a coat (CL)) are collected.
Therefore, there are 124(subjects) $\times$ 11(view angles) $\times$ 10 (6 NM + 2 BG + 2 CL)  = 13,640 gait sequences in the CASIA-B benchmark dataset. To evaluate the proposed method, we use the same protocol as \cite{chao2019gaitset}. Specifically, 124 subjects are divided into two parts: one is used as the training set to train the network and the other is taken as the test set. To be specific, 74 subjects for training, while the remaining 50 subjects are used for evaluation. During the training stage, all 10 groups of videos are used for training. In the test stage, four groups of videos (NM\#01-NM\#04) are used as the gallery set, and the remaining videos NM\#05-NM\#06, BG\#01-BG\#02 and CL\#01-CL\#02 are regarded as the probe set.

\noindent \textbf{OUMVLP} \cite{takemura2018multi} is one of the largest gait datasets. It includes 10,307 subjects and each subject contains two sets of gait sequences: the gallery set (Seq\#00) and probe set (Seq\#01). Each set is collected from 14 view angles ($0^\circ$-$90^\circ$ and $180^\circ$-$270^\circ$ with sampling interval $15^\circ$). We adopt the protocol \cite{chao2019gaitset, fan2020gaitpart} to evaluate the proposed method on the OUMVLP dataset. In the test stage, we take the sequences Seq\#01 as the gallery set, while the sequences Seq\#00 are used as the probe set.

\noindent \textbf{GREW} \cite{zhu2021gait} is a large-scale and in-the-wild dataset. It includes 26,345 subjects and 128,671 sequences. It also contains four data types: Silhouettes, Optical Flow, 2D Pose and 3D pose. The dataset has been divided into three parts, the training set, the validation set and the test set. The training set consists of 20,000 identities and 102,887 sequences, while the validation set contains 345 identities and 1,784 sequences. The test set includes 6,000 identities and each identity contains four sequences. In particular, GREW also contains a distractor set consisting of 233,857 sequences (unlabeled data). In the test stage, we take two sequences as the gallery set and the rest sequences are used as the probe set.

\noindent \textbf{Gait3D} \cite{zheng2022gait} is a large 3D representation-based gait recognition dataset. It contains 4,000 subjects and over 25,000 sequences. All sequences of Gait3D are collected in a large supermarket. To evaluate the performance of different algorithms, 4,000 subjects are divided into two sets: the training set (including 3000 subjects) and the test set (including 1000 subjects). For the test set, one sequence of each subject is selected as the probe set, while the rest sequences are regarded as the gallery set.

\subsection{Implementation Details}
To normalize the image size of gait silhouettes, we adopt the pre-processing method from \cite{chao2019gaitset}. For the CASIA-B, OUMVLP, and GREW datasets, the image size is normalized to $64 \times 44$. The image size of each frame on the Gait3D dataset is $128 \times 88$. The network structure on the CASIA-B dataset is shown in Fig.\ref{overview}. The number of global and local convolution layers for three stages is 1, 1 and 2, respectively. The output channels of ``stage 1'', ``stage 2'' and ``stage 3'' are 64, 128 and 128, respectively. Since OUMVLP, GREW and Gait3D datasets include more subjects than the CASIA-B dataset, we first add a spatial pooling layer after ``stage 3''. Then, we add a global and local convolution layer as ``stage 4'' to further extract gait features. The number of global and local convolution layers for the four stages is 2, 2, 2 and 4, respectively. The output channels of ``stage 1'', ``stage 2'', ``stage 3'' and ``stage 4'' are 64, 128, 256 and 512, respectively. In this setting, the local temporal aggregation is applied in the stage 2.
The parameter $m$ in Eqn. \ref{tri} is set to 0.2 and the parameter $p$ in Eqn. \ref{Equ_GeM} is initialized to 6.5. In Sec. \ref{Traintest}, we introduce the sampling strategy of our framework. The sampling strategy includes two parameters $P$ and $K$. For CASIA-B, the parameters $P$ and $K$ are set to 8 and 16, respectively. For OUMVLP, the parameters $P$ and $K$ are set to 32 and 8, respectively. Since each subject on the GREW and Gait3D datasets only contains a few sequences, $P \times K$ for both datasets is set to $32 \times 4$. During the training stage, the frame number of each batch is set to 30. In all experiments on CASIA-B, Adam is taken as the optimizer, while SGD is used as the optimizer for the other three datasets. The learning rate is initialized to 1e-4 for the experiments conducted on CASIA-B, and is set to 0.1 for the other three datasets. The iteration is set to 80k, 200k, 200k, and 150k for the CASIA-B, OUMVLP, GREW and Gait3D datasets, respectively.

\begin{table*}[htbp]
  \centering
  \caption{Rank-1 accuracy (\%) on OU-MVLP dataset under different view angles, excluding identical-view cases. The last eight rows show the results excluding invalid probe sequences.}
  \renewcommand\arraystretch{1.2} 
  \resizebox{0.99\textwidth}{!}{
    \begin{tabular}{cc|c|c|c|c|c|c|c|c|c|c|c|c|c|c}
    \toprule
    \multirow{2}[2]{*}{\textbf{Methods}} & \multicolumn{14}{|c|}{\textbf{Probe View}}                                                            & \multicolumn{1}{c}{\multirow{2}[2]{*}{\textbf{Mean}}}  \\
\cline{2-15}    \multicolumn{1}{c|}{} & $0^{\circ}$ & $15^{\circ}$ & $30^{\circ}$ & $45^{\circ}$ & $60^{\circ}$ & $75^{\circ}$ & $90^{\circ}$ & $180^{\circ}$ & $195^{\circ}$ & $210^{\circ}$ & $225^{\circ}$ & $240^{\circ}$ & $255^{\circ}$ & $270^{\circ}$ &   \\
    \midrule
    
    \multicolumn{1}{c|}{GEINet \cite{shiraga2016geinet}} & 23.2  & 38.1  & 48.0  & 51.8  & 47.5  & 48.1  & 43.8  & 27.3  & 37.9  & 46.8  & 49.9  & 45.9  & 45.7  & 41.0  & 42.5   \\
    \hline
    \multicolumn{1}{c|}{GaitSet \cite{chao2019gaitset}} & 79.3  & 87.9  & 90.0  & 90.1  & 88.0  & 88.7  & 87.7  & 81.8  & 86.5  & 89.0  & 89.2  & 87.2  & 87.6  & 86.2  & 87.1   \\
    \hline
    \multicolumn{1}{c|}{GaitPart \cite{fan2020gaitpart}} & 82.6  & 88.9  & 90.8  & 91.0  & 89.7  & 89.9  & 89.5  & 85.2  & 88.1  & 90.0  & 90.1  & 89.0  & 89.1  & 88.2  & 88.7   \\
    \hline
    \multicolumn{1}{c|}{GLN \cite{hou2020gait}}   & 83.8  & 90.0  & 91.0  & 91.2  & 90.3  & 90.0  & 89.4  & 85.3  & 89.1  & 90.5  & 90.6  & 89.6  & 89.3  & 88.5  & 89.2   \\
    \hline
    \multicolumn{1}{c|}{GaitKMM \cite{Zhang2021CVPR}} & 56.2  & 73.7  & 81.4  & 82.0  & 78.4  & 78.0  & 76.5  & 60.2  & 72.0  & 79.8  & 80.2  & 76.7  & 76.3  & 73.9  & 74.7   \\
    \hline
    \multicolumn{1}{c|}{SRN+CB \cite{hou2021set}} & 85.6  & 90.7  & 91.5  & 91.7  & 90.6  & 90.6  & 90.1  & 86.8  & 90.0  & 90.9  & 91.1  & 89.9  & 90.0  & 89.3  & 89.9   \\
    \hline
    \multicolumn{1}{c|}{CSTL \cite{huang2021context}} & \textbf{\textcolor{myblue}{87.1}}  & 91.0  & 91.5  & 91.8  & 90.6  & 90.8  & 90.6  & 89.4  & 90.2  & 90.5  & 90.7  & 89.8  & 90.0  & 89.4  & 90.2   \\
    \hline
    \multicolumn{1}{c|}{3D Local \cite{huang20213d}} & 86.1  & \textbf{\textcolor{myblue}{91.2}}  & \textbf{\textcolor{myblue}{92.6}}  & \textbf{\textcolor{myblue}{92.9}}  & \textbf{\textcolor{myblue}{92.2}}  & \textbf{\textcolor{myblue}{91.3}}  & \textbf{\textcolor{myblue}{91.1}}  & \textbf{\textcolor{myblue}{86.9}}  & \textbf{\textcolor{myblue}{90.8}}  & \textbf{\textcolor{myblue}{92.2}}  & \textbf{\textcolor{myblue}{92.3}}  & \textbf{\textcolor{myblue}{91.3}}  & \textbf{\textcolor{myblue}{91.1}}  & \textbf{\textcolor{myblue}{90.2}}  & \textbf{\textcolor{myblue}{90.9}}  \\
    \hline
    \multicolumn{1}{c|}{Our preliminary \cite{lin2020gait}} & 84.9  & 90.2  & 91.1  & 91.5  & 91.1  & 90.8  & 90.3  & 88.5  & 88.6  & 90.3  & 90.4  & 89.6  & 89.5  & 88.8  & 89.7   \\
    \hline
    \multicolumn{1}{c|}{Ours}  & \textbf{\textcolor{myred}{91.1}} & \textbf{\textcolor{myred}{92.6}} & \textbf{\textcolor{myred}{92.3}} & \textbf{\textcolor{myred}{92.5}} & \textbf{\textcolor{myred}{92.8}} & \textbf{\textcolor{myred}{92.2}} & \textbf{\textcolor{myred}{92.1}} & \textbf{\textcolor{myred}{92.4}} & \textbf{\textcolor{myred}{91.9}} & \textbf{\textcolor{myred}{91.7}} & \textbf{\textcolor{myred}{91.9}} & \textbf{\textcolor{myred}{92.1}} & \textbf{\textcolor{myred}{91.5}} & \textbf{\textcolor{myred}{91.4}} & \textbf{\textcolor{myred}{92.0}} \\

    \hline
    \\
    \hline
    \multicolumn{1}{c|}{GEINet \cite{shiraga2016geinet}} & 24.9  & 40.7  & 51.6  & 55.1  & 49.8  & 51.1  & 46.4  & 29.2  & 40.7  & 50.5  & 53.3  & 48.4  & 48.6  & 43.5  & 45.3   \\
    \hline
    \multicolumn{1}{c|}{GaitSet \cite{chao2019gaitset}} & 84.5  & 93.3  & 96.7  & 96.6  & 93.5  & 95.3  & 94.2  & 87.0  & 92.5  & 96.0  & 96.0  & 93.0  & 94.3  & 92.7  & 93.3   \\
    \hline
    \multicolumn{1}{c|}{GaitPart \cite{fan2020gaitpart}} & 88.0  & 94.7  & 97.7  & 97.6  & 95.5  & 96.6  & 96.2  & 90.6  & 94.2  & 97.2  & 97.1  & 95.1  & 96.0  & 95.0  & 95.1   \\
    \hline
    \multicolumn{1}{c|}{GLN \cite{hou2020gait}}   & 89.3  & 95.8  & 97.9  & 97.8  & 96.0  & 96.7  & 96.1  & 90.7  & 95.3  & 97.7  & 97.5  & 95.7  & 96.2  & 95.3  & 95.6   \\
    \hline
    \multicolumn{1}{c|}{SRN+CB \cite{hou2021set}} & \textbf{\textcolor{myblue}{91.2}}  & \textbf{\textcolor{myblue}{96.5}}  & \textbf{\textcolor{myblue}{98.3}}  & \textbf{\textcolor{myblue}{98.4}}  & 96.3  & 97.3  & 96.8  & 92.3  & \textbf{\textcolor{myblue}{96.3}}  & \textbf{\textcolor{myblue}{98.1}}  & \textbf{\textcolor{myblue}{98.1}}  & \textbf{\textcolor{myblue}{96.0}}  & \textbf{\textcolor{myblue}{97.0}}  & \textbf{\textcolor{myblue}{96.2}}  & 96.4   \\
    \hline

    \multicolumn{1}{c|}{3D Local \cite{huang20213d}} &  $-$     &     $-$   &     $-$   &      $-$  &     $-$   &  $-$      &     $-$   &      $-$  &     $-$   &     $-$   &    $-$    &    $-$    &    $-$    &     $-$   & \textbf{\textcolor{myblue}{96.5}}  \\
    \hline
    \multicolumn{1}{c|}{Our preliminary \cite{lin2021gait}} & 90.5  & 96.1  & 98.0  & 98.1  & \textbf{\textcolor{myblue}{97.0}}  & \textbf{\textcolor{myblue}{97.6}}  & \textbf{\textcolor{myblue}{97.1}}  & \textbf{\textcolor{myblue}{94.2}}  & 94.9  & 97.4  & 97.4  & 95.7  & 96.5  & 95.7  & 96.2   \\
    \hline
    \multicolumn{1}{c|}{Ours}  & \textbf{\textcolor{myred}{97.0}} & \textbf{\textcolor{myred}{98.6}} & \textbf{\textcolor{myred}{99.3}} & \textbf{\textcolor{myred}{99.3}} & \textbf{\textcolor{myred}{98.9}} & \textbf{\textcolor{myred}{99.2}} & \textbf{\textcolor{myred}{99.1}} & \textbf{\textcolor{myred}{98.3}} & \textbf{\textcolor{myred}{98.4}} & \textbf{\textcolor{myred}{99.0}} & \textbf{\textcolor{myred}{99.0}} & \textbf{\textcolor{myred}{98.6}} & \textbf{\textcolor{myred}{98.8}} & \textbf{\textcolor{myred}{98.6}} & \textbf{\textcolor{myred}{98.7}} \\

    \bottomrule
    \end{tabular}%
}
  \label{comparision_oumvlp}%
\end{table*}%

\subsection{Comparison with State-of-the-Art Methods}
In this section, we compare our method with state-of-the-art (SOTA) methods on four public datasets, CASIA-B \cite{yu2006framework}, OUMVLP \cite{takemura2018multi}, GREW \cite{zhu2021gait} and Gait3D \cite{zheng2022gait}.

\noindent \textbf{Evaluation on CASIA-B.} 
The experimental results are shown in Table. \ref{comparision_casia}. The compared methods include GaitSet \cite{chao2019gaitset}, GaitPart \cite{fan2020gaitpart}, MT3D \cite{lin2020gait}, 3D Local \cite{huang20213d}, CSTL \cite{huang2021context} and our preliminary \cite{lin2021gait}. It can be observed that the proposed method achieves appealing performance compared with other SOTA methods. 
Furthermore, we analyze the performance of different methods under different carry conditions (NM, BG and CL). As shown in Table. \ref{comparision_casia}, the accuracy of CSTL in the settings of NM, BG and CL is 97.8\%, 93.6\% and 84.2\%, respectively. For the proposed method, the accuracy in NM, BG and CL settings is 98.0\%, 95.4\% and 87.3\%, which outperforms CSTL by 0.2\%, 1.8\% and 3.1\%, respectively. It can be observed that our method significantly improves the performance under complex conditions (e.g., BG and CL). This is because the proposed method takes full advantage of both global visual information and local posture details. However, CSTL pays more attention to the global contextual information but neglects the local changes in gait sequences. 
Moreover, we also compare GaitGL with our previous work trained by fixed masks. As illustrated in Table. \ref{comparision_casia}, the accuracy of our previous work is 97.4\%, 94.5\% and 83.6\%, respectively. It can be found that the network trained by masked-based strategies outperforms our previous work by 0.6\% in the NM setting, 0.9\% in the BG setting and 3.7\% in the CL setting, respectively. This is because that the proposed masked-based strategies can better utilize local posture details and thus increases the robustness of feature representations.

\noindent \textbf{Evaluation on OUMVLP.} 
For the OUMVLP dataset, we compare the proposed method with several outstanding gait recognition methods, including GEINet\cite{shiraga2016geinet}, GaitSet\cite{chao2019gaitset}, GaitPart\cite{fan2020gaitpart}, GLN\cite{hou2020gait}, GaitKMM\cite{Zhang2021CVPR}, SRN+CB\cite{hou2021set}, CSTL\cite{huang2021context}, 3D Local \cite{huang20213d} and Our preliminary \cite{lin2021gait}. We take the same protocol as the  GaitSet and GaitPart methods for fair comparison. The experimental results are shown in Table. \ref{comparision_oumvlp}. We can observe that the average accuracy of the proposed method is 98.7\%, which outperforms 3D Local by 2.2\%. Moreover, it can be found that our method achieves higher performance improvement on the horizontal views (e.g., $0^{\circ}$and $180^{\circ}$) than other views. For example, the accuracy of 3D Local on $0^{\circ}$ and $180^{\circ}$ views is only 86.1\% and 86.9\%, respectively. They are much lower than the average accuracy of SRN+CB (90.9\%). This is because the views $0^{\circ}$ and $180^{\circ}$ contain less information than other views. Thus, it is hard to extract discriminative features from them. 
In contrast, the accuracy of our method on $0^{\circ}$ and $180^{\circ}$ views is 91.1\% and 92.4\%, respectively. They outperform the SRN+CB by 5.0\% and 5.5\%, respectively. This is because our method generates global and local feature representations to portray human gaits. The local feature representations can better utilize local posture details from limited views.

\begin{table}[htbp]
  \centering
  \caption{Rank-1 accuracy (\%), Rank-5 accuracy (\%), Rank-10 accuracy (\%), and Rank-20 accuracy (\%) on the GREW dataset.}
  \renewcommand\arraystretch{1.2} 
    \begin{tabular}{c|c|c|c|c}
    \toprule
    Methods & \multicolumn{1}{c|}{Rank-1} & \multicolumn{1}{c|}{Rank-5} & \multicolumn{1}{c|}{Rank-10} & \multicolumn{1}{c}{Rank-20} \\
    \midrule
    GEINet \cite{shiraga2016geinet} & 6.8   & 13.4  & 17.0  & 21.0  \\
    \hline
    TS-CNN \cite{wu2016comprehensive} & 13.6  & 24.6  & 30.2  & 37.0  \\
    \hline
    GaitSet \cite{chao2019gaitset} & 46.3  & 63.6  & 70.3  & 76.8  \\
    \hline
    GaitPart \cite{fan2020gaitpart} & 44.0  & 60.7  & 67.3  & 73.5  \\
    \hline
    CSTL \cite{huang2021context} & \textbf{\textcolor{myblue}{50.6}}  & \textbf{\textcolor{myblue}{65.9}}  & \textbf{\textcolor{myblue}{71.9}}  & \textbf{\textcolor{myblue}{76.9}}  \\
    \hline
    Ours  & \textbf{\textcolor{myred}{68.0}} & \textbf{\textcolor{myred}{80.7}} & \textbf{\textcolor{myred}{85.0}} & \textbf{\textcolor{myred}{88.2}} \\
    \bottomrule
    \end{tabular}%
  \label{comparision_GREW}%
\end{table}%

\begin{table}[htbp]
  \centering
  \caption{Rank-1 accuracy (\%), Rank-5 accuracy (\%), mAP (\%) and mINP on the Gait3D dataset.}
    \renewcommand\arraystretch{1.2} 
    \begin{tabular}{c|c|c|c|c}
    \toprule
    Methods & \multicolumn{1}{c|}{Rank-1} & \multicolumn{1}{c|}{Rank-5} & \multicolumn{1}{c|}{mAP} & \multicolumn{1}{c}{mINP}  \\
    \midrule
    PoseGait \cite{liao2020model} & 0.24  & 1.08  & 0.47  & 0.34  \\
    \hline
    GaitGraph \cite{teepe2021gaitgraph} & 6.25  & 16.23 & 5.18  & 2.42  \\
    \hline
    GEINet \cite{shiraga2016geinet} & 7.00  & 16.30  & 6.05  & 3.77   \\
    \hline
    GaitSet \cite{chao2019gaitset} & 42.60  & 63.10  & 33.69  & 19.69   \\
    \hline
    GaitPart \cite{fan2020gaitpart} & 29.90  & 50.60  & 23.34  & 13.15   \\
    \hline
    GLN \cite{hou2020gait}  & 42.20  & 64.50  & 33.14  & 19.56   \\
    \hline
    CSTL \cite{huang2021context}  & 12.20  & 21.70  & 6.44  & 3.28   \\
    \hline
    SMPLGait \cite{zheng2022gait} & \textbf{\textcolor{myblue}{53.20}}  & \textbf{\textcolor{myblue}{71.00}}  & \textbf{\textcolor{myblue}{42.43}}  & \textbf{\textcolor{myblue}{25.97}}   \\
    \hline
    Ours  &  \textbf{\textcolor{myred}{63.80}}     &   \textbf{\textcolor{myred}{80.50}}    &   \textbf{\textcolor{myred}{55.88}}    &  \textbf{\textcolor{myred}{36.71}}  \\
    \bottomrule
    \end{tabular}%
  \label{comparision_Gait3D}%
\end{table}%

\noindent \textbf{Evaluation on the GREW and Gait3D datasets.} 
Since the CASIA-B and OUMVLP datasets are collected from an experimental environment, we further conduct several experiments on some in-the-wild datasets, e.g., GREW and Gait3D. Both datasets are collected in the wild. Thus, they contain more complex conditions, such as occlusions of human bodies and misaligned postures, than the CASIA-B and OUMVLP datasets. 
We follow the same settings \cite{zhu2021gait,zheng2022gait} to conduct our experiments. The experimental results are illustrated in Table. \ref{comparision_GREW} and Table. \ref{comparision_Gait3D}, respectively. The compared methods include GEINet \cite{shiraga2016geinet}, TS-CNN \cite{wu2016comprehensive}, GaitSet \cite{chao2019gaitset}, GaitPart \cite{fan2020gaitpart}, CSTL \cite{huang2021context}, PoseGait \cite{liao2020model}, GaitGraph \cite{teepe2021gaitgraph}, GLN \cite{hou2020gait} and SMPLGait \cite{zheng2022gait}.
It can be observed that our method achieves appealing performance on both datasets. Moreover, we can observe that the accuracy of local-based methods, such as GaitPart, is lower than those global-based methods, such as Gaitset. This is because the data collected from the real-world scenes contain numerous noises. Since local-based methods pay more attention to local posture details, they are more likely to over-fit these noises. However, our method can adaptively learn and fuse global visual information and local posture details. As a result, it can generate more discriminative feature representations and thus improves its performance on in-the-wild datasets.

\begin{figure*}[htbp]

\subfigure[Horizontal masks at the part level]{
\centering
\includegraphics[width=0.33\textwidth]{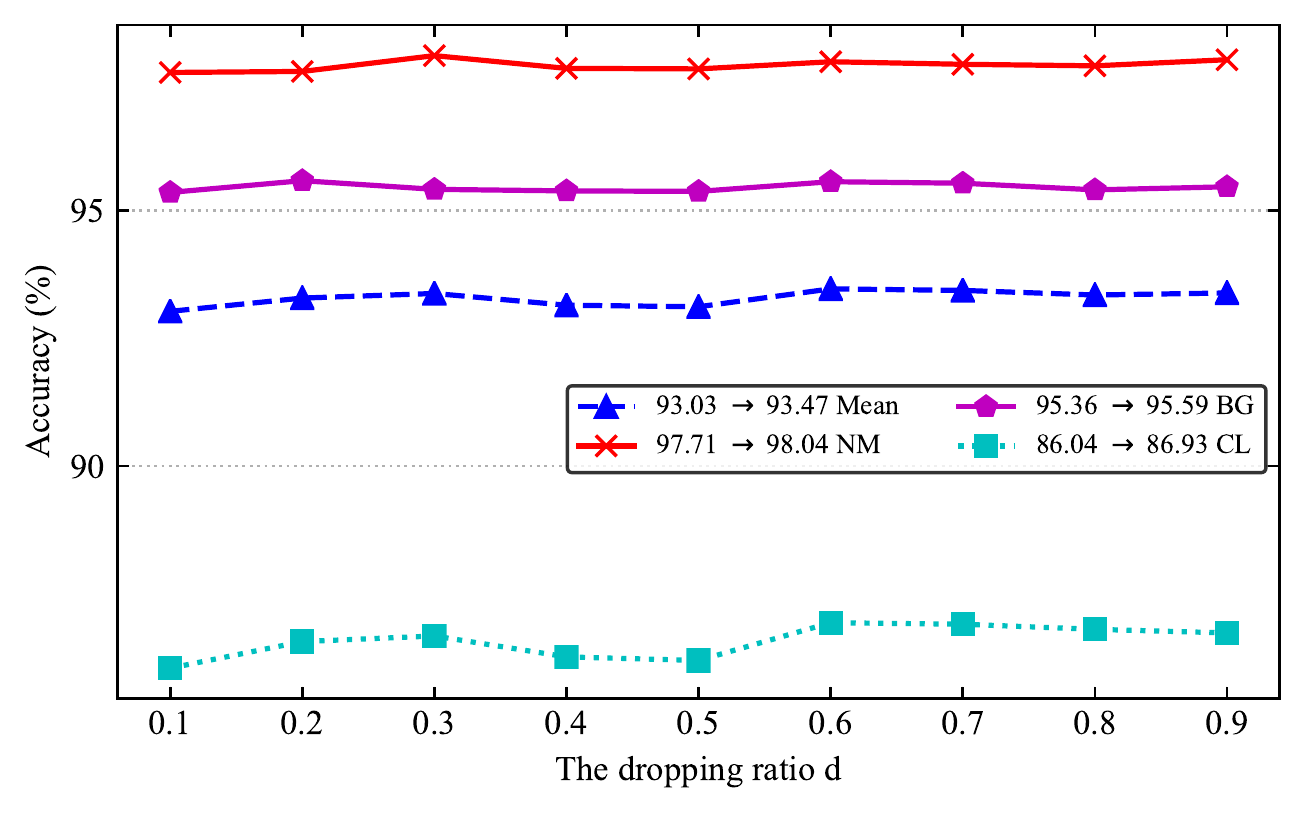}
}
\subfigure[Vertical masks at the part level]{
\centering
\includegraphics[width=0.33\textwidth]{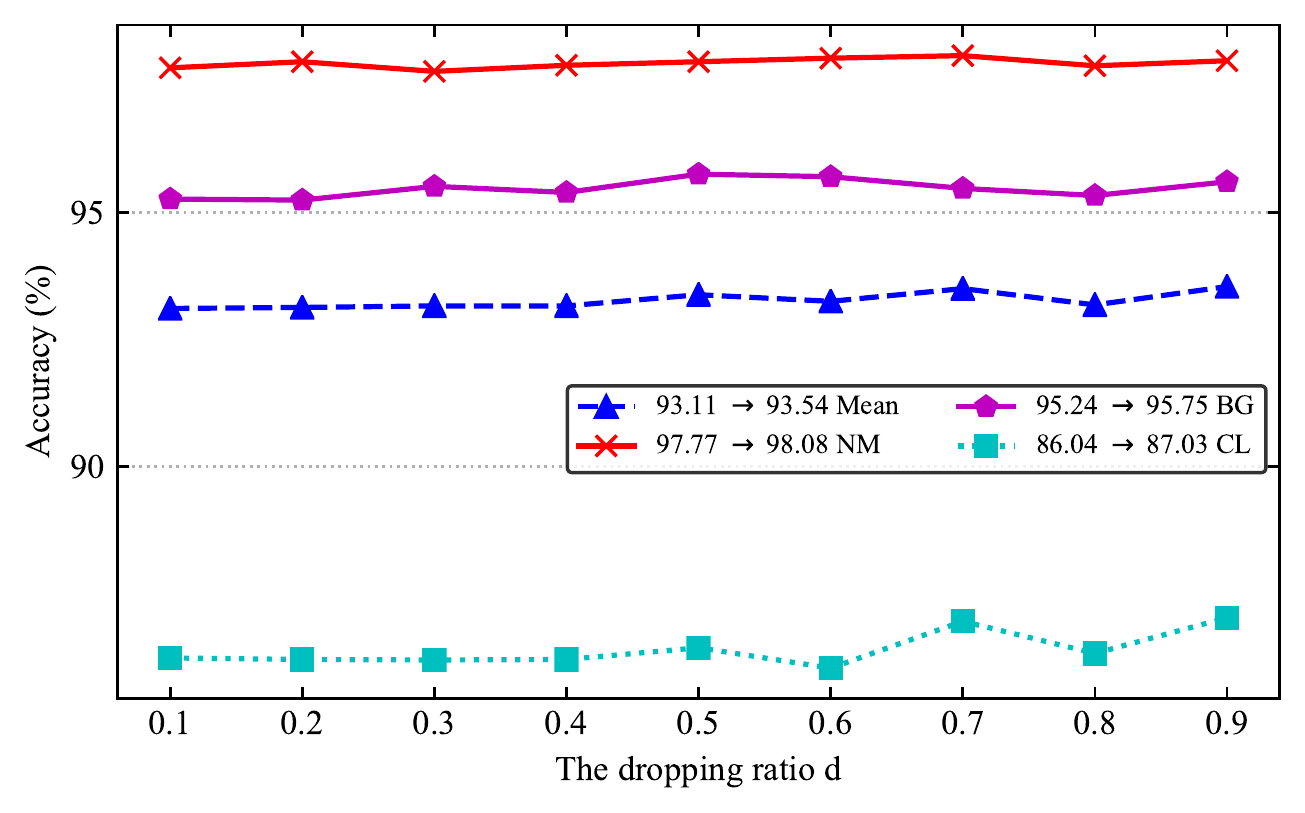}
}
\subfigure[Horizontal masks at the strip level]{
\centering
\includegraphics[width=0.33\textwidth]{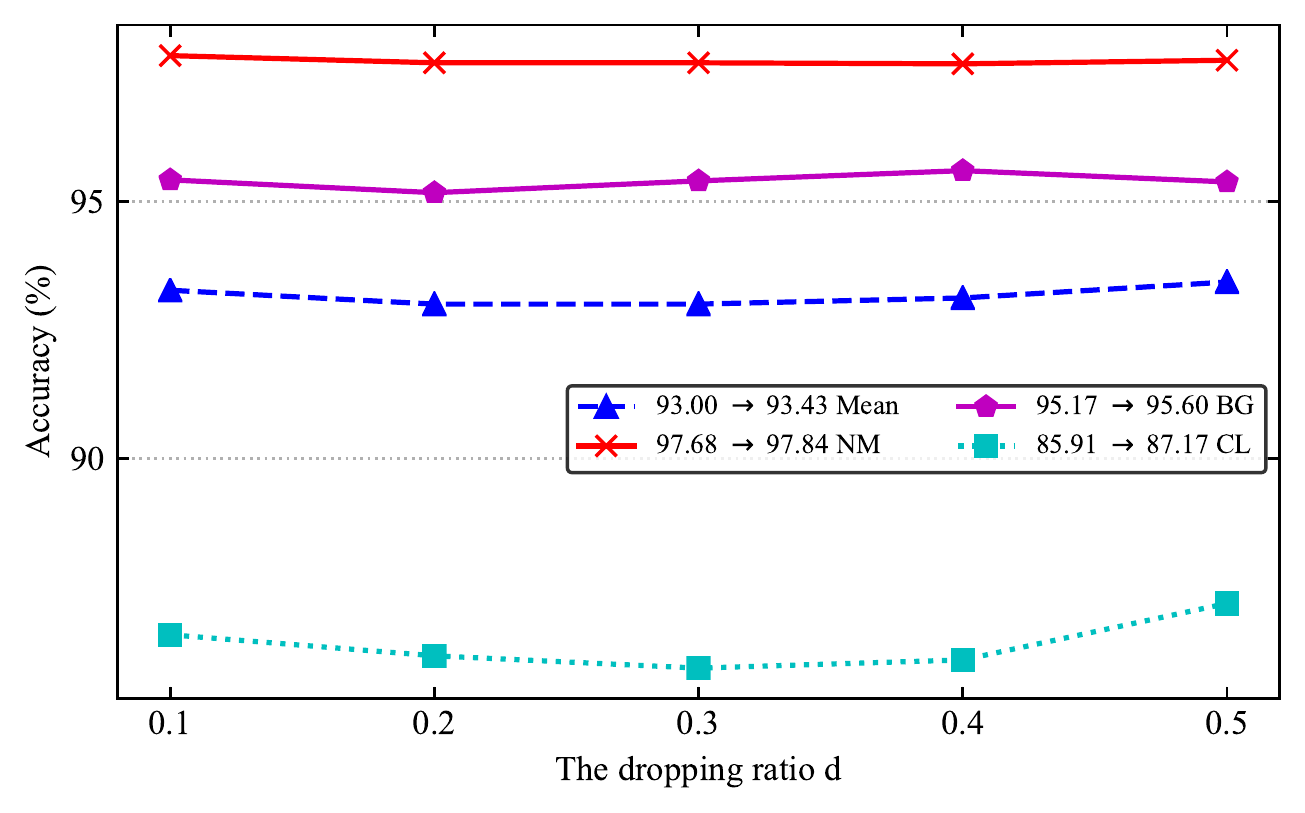}
}

\subfigure[Vertical masks at the strip level]{
\centering
\includegraphics[width=0.33\textwidth]{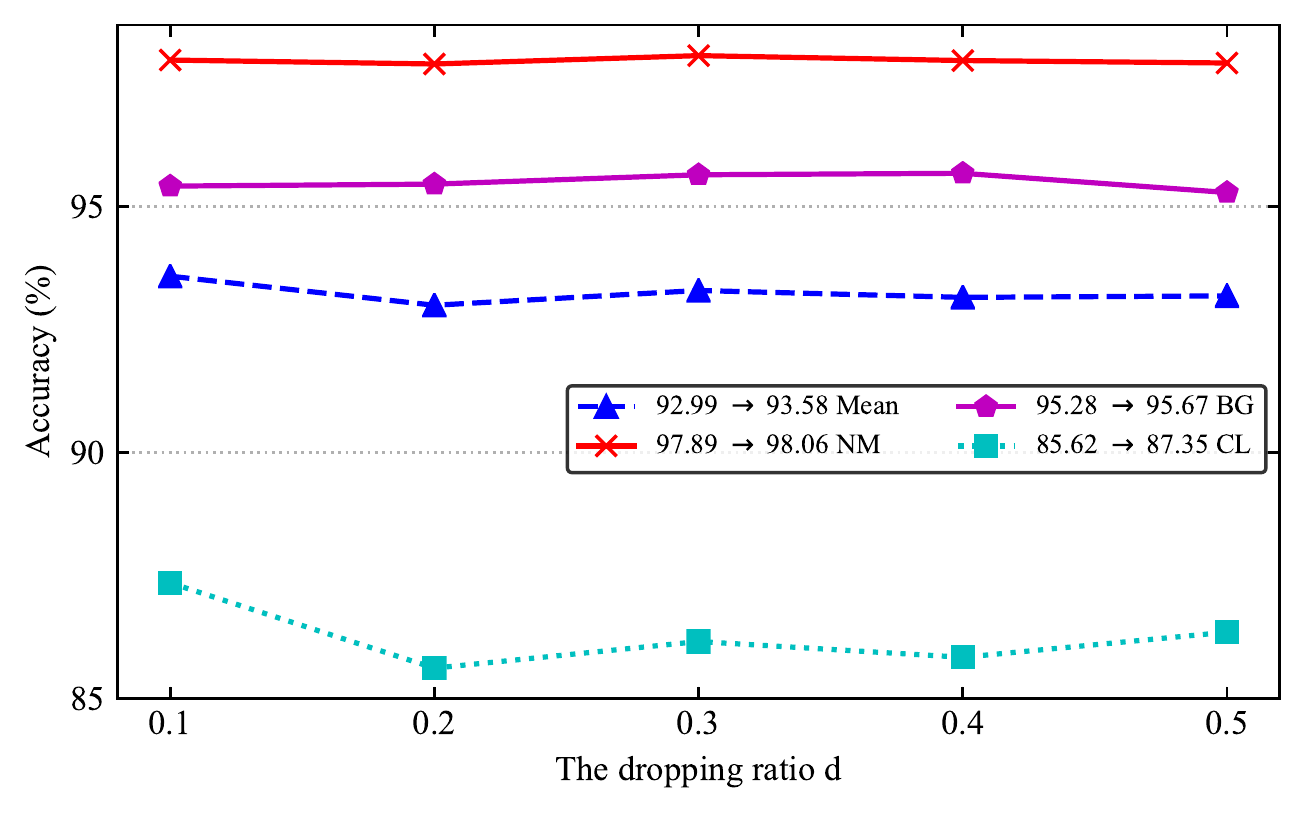}
}
\centering
\subfigure[Pixel-level masks]{
\centering
\includegraphics[width=0.33\textwidth]{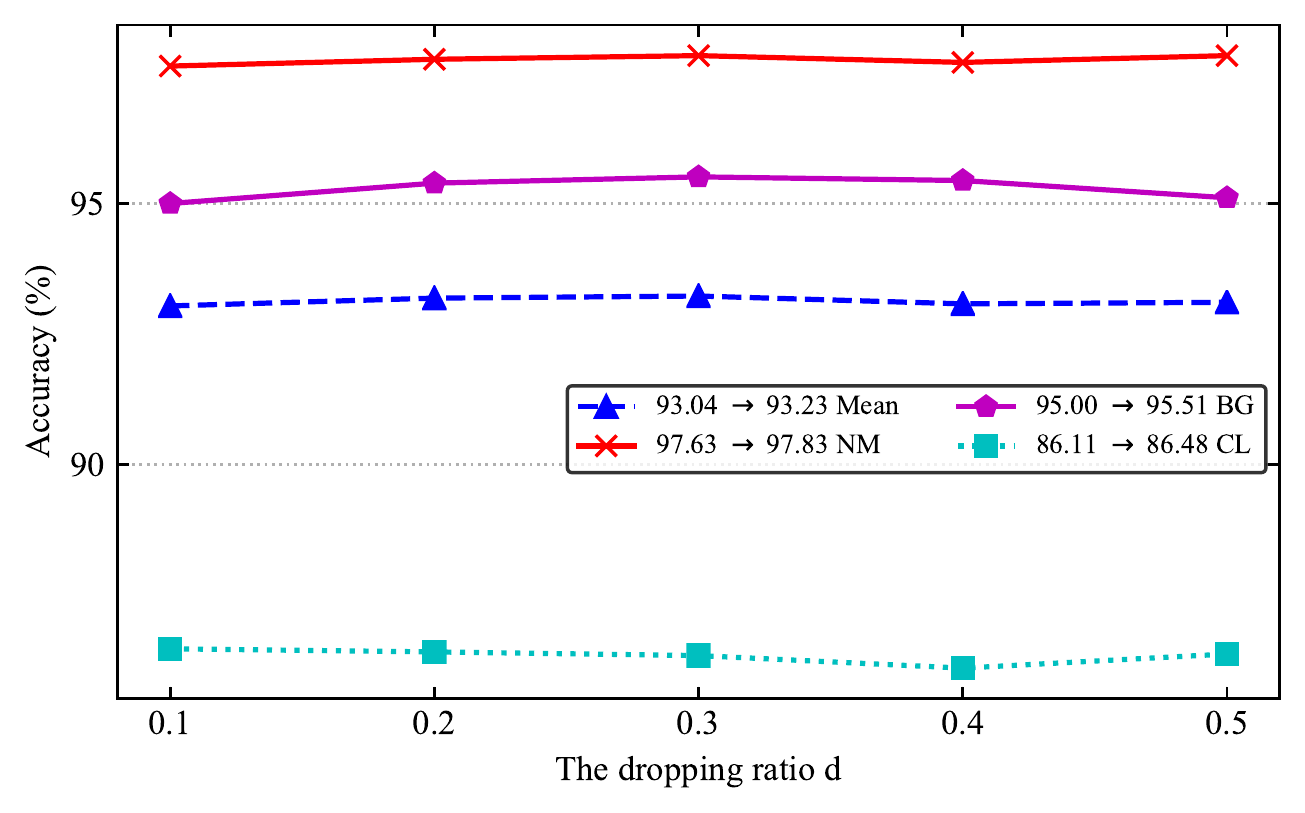}
}

\label{fig_exp_masks}
\caption{Rank-1 accuracy (\%) of different mask strategies with different dropping ratios. The red line, purple line, cyan line mean the recognition accuracy under the settings of normal walking, walking with a bag and walking with a coat, respectively. The blue line represents an average accuracy for normal walking, walking with a bag and walking with a coat.}
\end{figure*}

\subsection{Ablation Study} \label{Ablation_Study}
In this paper, we present several key modules, including Local Temporal Aggregation (LTA) and Global and Local Convolutional Layer (GLCL). To verify the effectiveness of the proposed key modules, we conduct several ablation studies in the following parts.

\noindent \textbf{Analysis of global and local convolutional layer.}  
In this paper, we propose a novel Global and Local Convolutional Layer to generate more discriminative representations. To verify the effectiveness of the proposed GLCL, we conduct some ablation studies on the CASIA-B dataset. The experimental results are shown in Table. \ref{TAB_GLFE}. It can be observed that the average accuracy using the proposed GLCL is 93.6\%, which outperforms the global-based method and the local-based method by 1.3\% and 1.0\%, respectively. This is because the global feature extractor is hard to utilize local details in deep convolution layers, while the local feature extractor neglects the relation of different local regions. In contrast, the proposed GLCL effectively combines both global contextual information and local detailed information and thus improves the discriminativeness of feature representations.

\noindent \textbf{Analysis of different mask strategies.}  
We also conduct some ablation studies to analyze the performance impact of different mask strategies. Table. \ref{TAB_GLFE} shows that the recognition accuracy of three random mask strategies is about 93.5\%, which outperforms the fixed mask strategy by 0.6\%. This is because the proposed mask strategies generate various local feature maps to train the local feature extractor. Thus, the local feature extractor can effectively utilize the local information of feature maps. On the other hand, we also observe that the method with the strip-level masks at the vertical axis achieves better performance than other mask strategies. Therefore, we finally choose it to implement our model.

\noindent \textbf{Analysis of different mask ratios.} 
To better utilize the information of local posture details, we propose a novel mask-based local feature extractor. As shown in Fig. \ref{Fig_diff_region}, it includes three different mask strategies: Part-level Mask, Strip-level Mask and Pixel-level Mask. As described in Sec. \ref{sec_LFR}, each mask strategy contains a dropping ratio parameter. In this section, we conduct several experiments to analyze the impact of different dropping ratios. To be specific, the parameter range of horizontal and vertical masks at the part level is from 0.1 to 0.9. The parameter range of strip-level masks and pixel-level masks is from 0.1 to 0.5. The experimental results are shown in Fig. \ref{fig_exp_masks}. It can be observed that the accuracy curve of a mask strategy with different dropping ratios is stable. That means that our method is insensitive to the settings of the dropping ratio. This is because the proposed three mask strategies generate pairs of complementary masks to occlude gait feature maps. Then, we extract the local features of each occluded feature map. After that, the occluded feature maps are fused into a feature map. The fused map does not lose any regions' information. Therefore, the generated feature representations are robust to a mask strategy with different dropping ratios. As a result, we do not need to carefully set the dropping ratio on different datasets.

\begin{table}[t]
  \centering
  \renewcommand\arraystretch{1.2} 
  \caption{Rank-1 accuracy (\%) of different mask strategies on the CASIA-B dataset. ``H'' indicates horizontal masks, while ``V'' means vertical masks.}
    \begin{tabular}{c|c|c|c|c|c}
    \toprule
    \multicolumn{2}{c|}{Methods} & \multirow{2}[2]{*}{NM} & \multirow{2}[2]{*}{BG} & \multirow{2}[2]{*}{CL} & \multirow{2}[2]{*}{Mean}  \\
\cline{1-2}    \multicolumn{1}{c|}{Global} & \multicolumn{1}{c|}{Local} &       &       &       &   \\
    \midrule
    \checkmark &       &    97.4   &   94.5    &   85.1    & 92.3  \\
    \hline
          & Strip-level masks (V) &    97.3   &    94.8   &    85.8   &  92.6 \\
    \hline
    \checkmark & Part-level fixed masks & 97.7  & 95.0  & 86.0  & 92.9   \\
    \hline
    \checkmark & Part-level masks (H) & 97.9  & 95.6  & 86.9  & 93.5   \\
    \hline
    \checkmark & Part-level masks (V) & 98.0  & \textbf{95.6} & 87.0  & 93.5   \\
    \hline
    \checkmark & Strip-level masks (H) & 97.7  & 95.4  & 87.2  & 93.4   \\
    \hline
    \checkmark & Strip-level masks (V) & \textbf{98.0} & 95.4  & \textbf{87.3} & \textbf{93.6}  \\
    \hline
    \checkmark & Pixel-level masks & 97.8  & 95.5  & 86.3  & 93.2   \\
    \bottomrule
    \end{tabular}%
  \label{TAB_GLFE}%
\end{table}%

\begin{table}[t]
  \centering
  \renewcommand\arraystretch{1.8} 
  \caption{Accuracy ($\%$) of different spatial feature mapping. $F_{Max}^{1\times1\times W_{2}}$ and $F_{Avg}^{1\times1\times W_{2}}$ indicate the average pooling layer and the max pooling layer, respectively. $GeM$ means the GeM pooling layer.}
  \resizebox{0.48\textwidth}{!}{
    \begin{tabular}{c|c|c|c|c|c|c}
    \toprule
    \multicolumn{3}{c|}{\textbf{Spatial Feature Mapping}} & \multirow{2}[2]{*}{NM} & \multirow{2}[2]{*}{BG} & \multirow{2}[2]{*}{CL} & \multirow{2}[2]{*}{Mean} \\
\cline{1-3}    $F_{Max}^{1\times1\times1\times W_{fm}}$   & $F_{Avg}^{1\times1\times1\times W_{fm}}$   & $F_{GeM}$   &       &       &       &  \\
    \midrule
    \checkmark      &       &       & 97.8  & 94.8  & 85.4  & 92.7   \\
    \hline
          & \checkmark      &       & 97.2  & 94.6  & 84.9  & 92.2    \\
    \hline
    \checkmark      & \checkmark      &       & 97.7  & 94.7  & 85.1  & 92.5     \\
    \hline
          &       & \checkmark        & \textbf{98.0}  & \textbf{95.4}  & \textbf{87.3} & \textbf{93.6}  \\
    \bottomrule
    \end{tabular}%
  } 
  \label{comparion_spm}%
\end{table}%

\begin{figure*}[htbp]

\subfigure[Global feature representations]{
\centering
\includegraphics[width=0.33\textwidth]{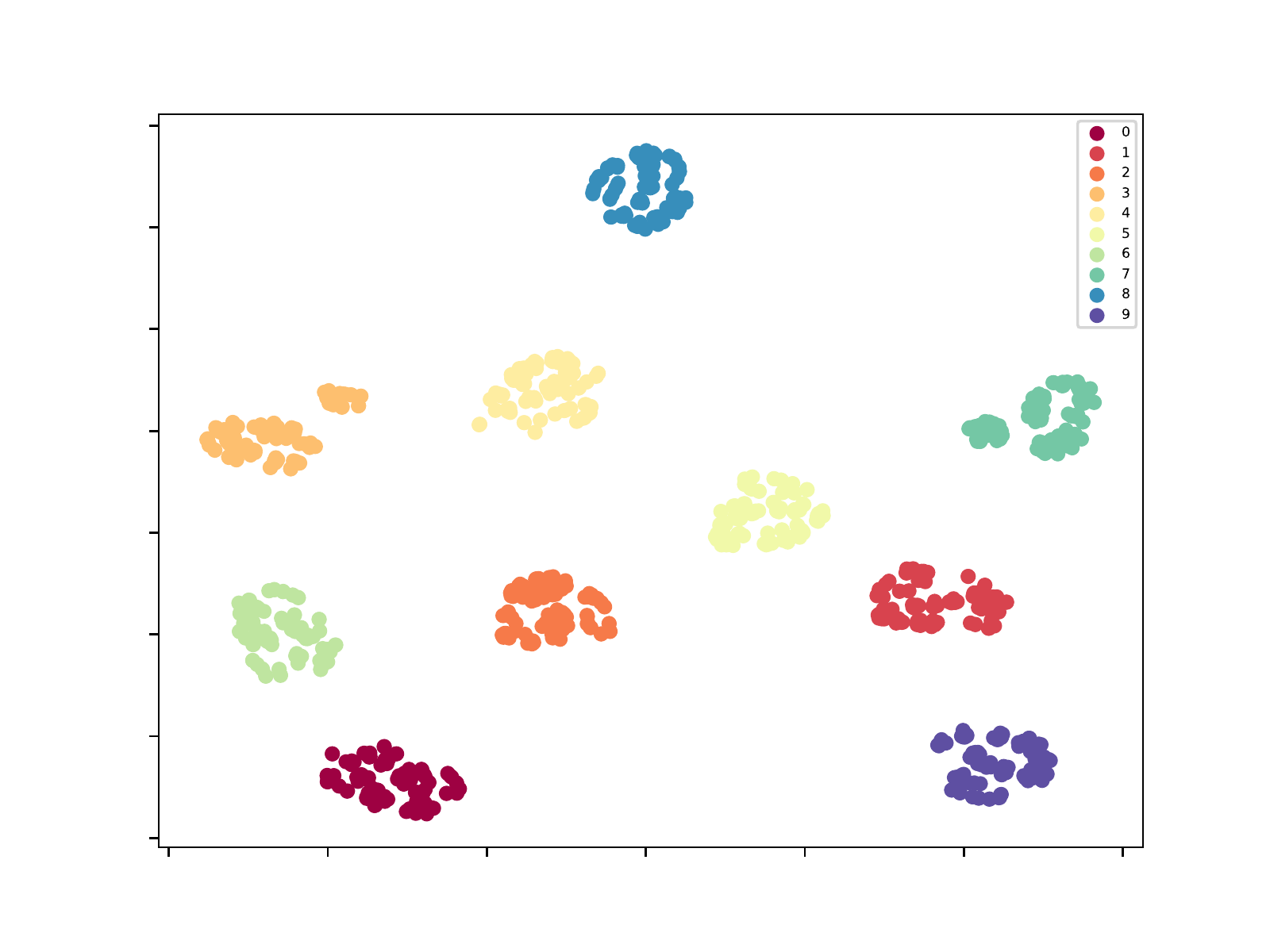}
}
\subfigure[Local feature representations]{
\centering
\includegraphics[width=0.33\textwidth]{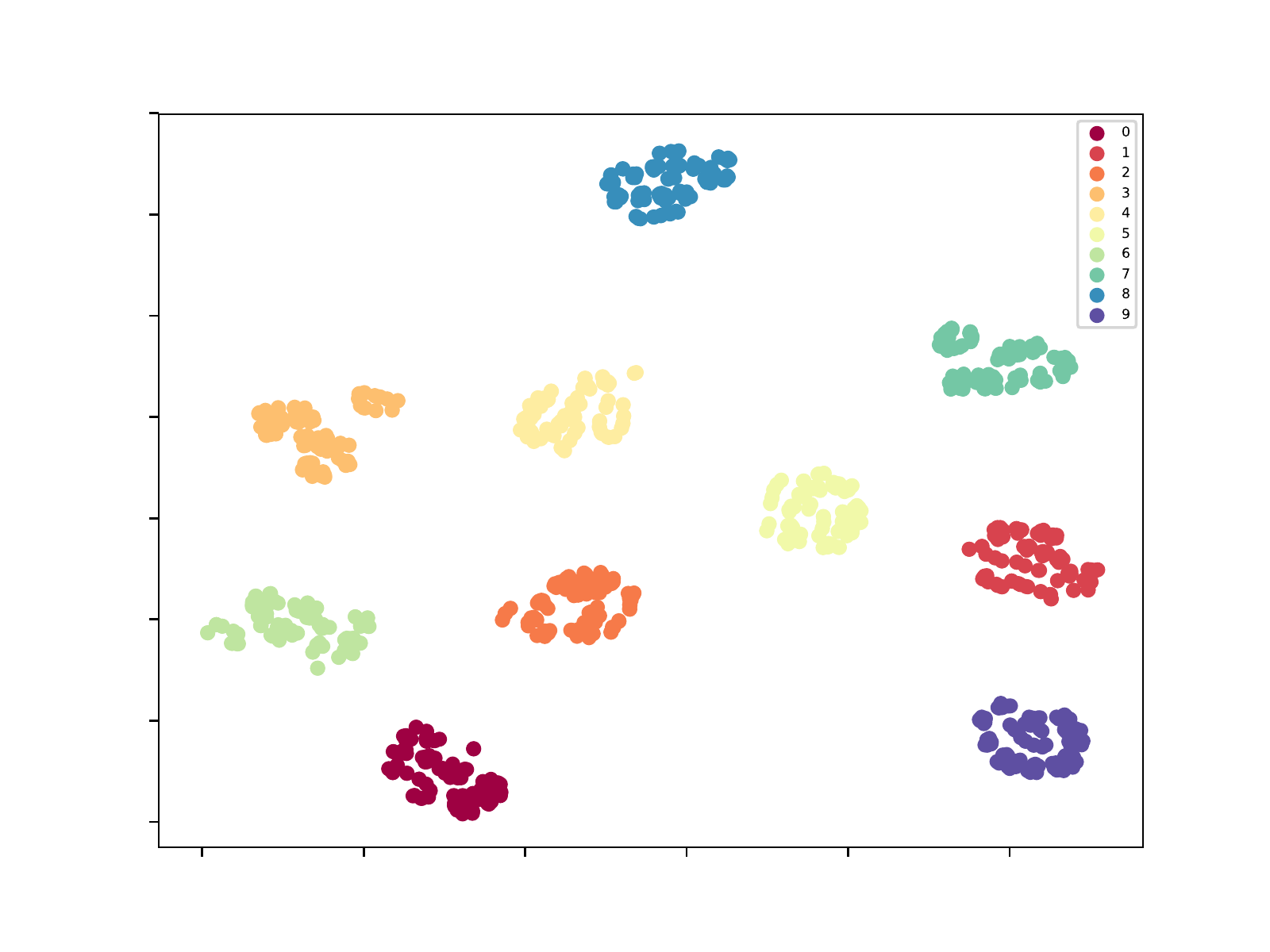}
}
\subfigure[Global and local feature representations]{
\centering
\includegraphics[width=0.33\textwidth]{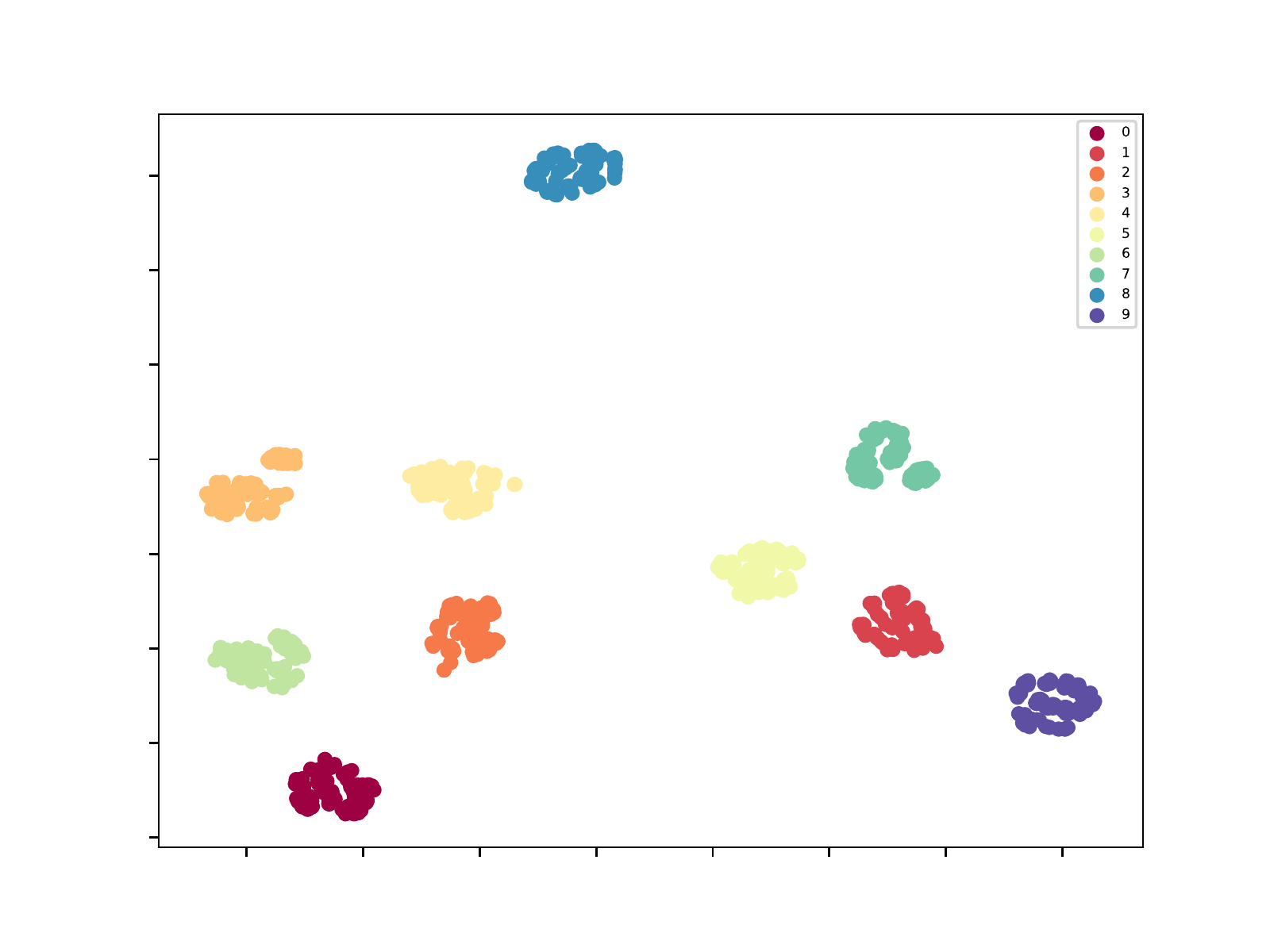}
}
\vspace*{-1em}
\caption{t-SNE visualization examples of the global feature representations, local feature representations and global and local feature representations on CASIA-B test dataset. We visualize 10 identities, each of which is selected 20 samples. Each point represents a sample and each color defines a class.
}
\label{Fig_vis}
\end{figure*}

\noindent \textbf{Analysis of spatial feature mapping.} 
Traditional methods always use a combined pooling layer, consisting of a max pooling operation and an average pooling operation, to aggregate the spatial information as shown in Eqn. \ref{Equ_MA}. In this paper, we propose the GeM pooling layer to adaptively integrate the spatial information of feature maps. To verify the effectiveness of the GeM pooling layer, we conduct several experiments with different feature mapping strategies on the CASIA-B dataset. The experimental results are illustrated in Table. \ref{comparion_spm}. We can observe that our method with the GeM pooling layer achieves better performance than other strategies in all conditions. To be specific, the average accuracy of using only $F_{Max}(\cdot)$ or only $F_{Avg}(\cdot)$ is 92.7\% and 92.2\%, respectively. The combination of using both $F_{Max}(\cdot)$ and $F_{Avg}(\cdot)$ is 92.5\%. However, our method with the GeM pooling layer achieves an accuracy of 93.6\%, which outperforms other strategies by 0.9\%. This is because the GeM pooling layer introduces a learnable parameter and thus can better fuse the spatial information of feature maps.

\begin{table}[h]
  \centering
  \caption{Accuracy($\%$) of different pooling layer combinations. SP indicates the spatial pooling layer and LTA means local temporal aggregation.}
  \renewcommand\arraystretch{1.2} 
  \resizebox{0.48\textwidth}{!}{
    \begin{tabular}{c|c|c|c|c|c}
    \toprule

    1st pooling layer & 2nd pooling layer &    NM   &   BG    &  CL & Mean\\
    \midrule
    SP    & SP    & 96.5  & 91.4  & 78.2  & 88.7 \\
    \hline
    SP    & LTA   &  97.5 & 94.5 & 84.2 & 92.2 \\
    \hline
    LTA   & SP    & \textbf{98.0}  & \textbf{95.4}  & \textbf{87.3} & \textbf{93.6}  \\
    \hline
    LTA   & LTA   & 96.3  & 92.3  & 80.5 & 89.7 \\
    \bottomrule
    \end{tabular}%
  } 
  \label{tab_Downsampling}%
\end{table}%

\noindent \textbf{Analysis of local temporal aggregation.}
In this paper, we propose a local temporal aggregation (LTA) operation to trade off temporal and spatial information of feature maps. To analyze the contribution of the LTA operation, we conduct several ablation studies with different combinations of pooling layers. The experiments are all implemented on the CASIA-B dataset and are shown in Table. \ref{tab_Downsampling}. 
It is well known that spatial pooling (SP) may lead to spatial information loss, so only using SP layers degrades the representation ability. Based on our observation that the adjacent frames are similar, we propose a novel LTA operation to mitigate the spatial information loss in SP. LTA reduces redundant temporal information by aggregation while fully exploiting the spatial information. 
However, if only LTA modules are used, the temporal information may be lost, and recognition performance will decrease. As indicated by Table \ref{tab_Downsampling}, ``LTA+SP” better balances the leverage of temporal and spatial information, and outperforms ``SP+SP” and ``LTA+LTA”.

\subsection{Visualization}
In this section, we use the t-SNE \cite{van2008visualizing} technique to visualize the feature distributions of different feature representations, including global feature representations, local feature representations and global and local feature representations. The visualization is illustrated in Fig. \ref{Fig_vis}. It can be observed that the feature distributions of using only global feature representations or local feature representations are incompact. In other words, the intra-class distance of both representations is too large. Thus, it is difficult for them to identify some samples that are far from the class center.
However, for the proposed global and local feature representations, the feature distributions of different samples from the same class are close. Therefore, our method is more likely to correctly identify some hard samples. The visualization result further proves the effectiveness of our method.

\section{Conclusion}
In this paper, we propose a novel gait recognition framework, dubbed GaitGL, to generate more comprehensive feature representations. To be specific, GaitGL is built based on multiple Global and Local Convolutional layers, consisting of a GFR extractor and a mask-based LFR extractor. The GFR extractor is presented to take advantage of global visual information, while the mask-based LFR extractor aims to utilize the local posture details of gait sequences. Furthermore, we propose a novel mask-based partition strategy to generate various local feature maps that can be used to train the network effectively. The experimental results on CAISA-B, OUMVLP, GREW and Gait3D datasets demonstrate that our method achieves appealing performance compared with SOTA methods. Meanwhile, extensive ablation studies show the effectiveness of the proposed key modules.


%



\ifCLASSOPTIONcompsoc
  \section*{Acknowledgments}
\else
  \section*{Acknowledgment}
\fi
The authors would like to thank the anonymous reviewers and the associate editor for their helpful suggestions and valuable comments.
This work was supported by the National Natural Science Foundation of China (61976017 and 61601021), the Beijing Natural Science Foundation (4202056), the Fundamental Research Funds for the Central Universities (2022JBMC013) and the Australian Research Council (DP220100800). The support and resources from the Center for High Performance Computing at Beijing Jiaotong University (http://hpc.bjtu.edu.cn) are gratefully acknowledged.

\ifCLASSOPTIONcaptionsoff
  \newpage
\fi



\bibliographystyle{IEEEtran}
\bibliography{egbib}

\end{document}